\pdfoutput=1

\documentclass[11pt]{article}

\usepackage[preprint]{acl}

\usepackage{times}
\usepackage{latexsym}

\usepackage[T1]{fontenc}

\usepackage[utf8]{inputenc}

\usepackage{microtype}

\usepackage{inconsolata}

\usepackage{graphicx}

\usepackage{xspace}

\usepackage{multirow}
\usepackage{enumitem}
\usepackage{booktabs}
\usepackage{adjustbox}
\usepackage{amsmath}
\usepackage{booktabs}
\usepackage{array}
\usepackage{longtable}
\usepackage{amssymb}

\newcommand{\camerareadytext}[1]{}

\usepackage{listings}
\usepackage{xcolor}
\usepackage{xcolor,soul}

\definecolor{mygreen}{HTML}{74c476}
\definecolor{myred}{HTML}{f4a5a5}
\definecolor{myyellow}{HTML}{f8e19f}
\definecolor{myblue}{HTML}{6495ed}

\newcommand{\fref}[1]{Figure~\ref{#1}}
\newcommand{\tref}[1]{Table~\ref{#1}}

\title{Cross-Lingual Prompt Steerability: Towards Accurate and Robust LLM Behavior across Languages}

\author{
  Lechen Zhang$^{\ddagger}$\thanks{~~Equal contribution}  ~ Yusheng Zhou$^{\dagger}$\footnotemark[1] ~ Tolga Ergen$^{\sharp}$ \\ 
  \textbf{Lajanugen Logeswaran}$^\sharp$ ~
  \textbf{Moontae Lee}$^{\sharp \diamond}$ ~ \textbf{David Jurgens}$^{\dagger}$ \\
$^\ddagger$University of Illinois Urbana-Champaign  ~
$^\dagger$University of Michigan  \\
$^{\diamond}$University of Illinois Chicago ~
$^\sharp$LG AI Research  \\
$^{\ddagger}${ \tt \{lechenz3\}@illinois.edu} ~
$^{\dagger}${ \tt \{yszhou, jurgens\}@umich.edu}\\
~ $^\sharp${ \tt \{tergen, llajan, moontae.lee\}@lgresearch.ai}
}

\begin{document}
\maketitle
\begin{abstract}
System prompts provide a lightweight yet powerful mechanism for conditioning large language models (LLMs) at inference time. While prior work has focused on English‑only settings, real‑world deployments benefit from having a single prompt to operate reliably across languages. This paper presents a comprehensive study of how different system prompts steer models toward accurate and robust cross-lingual behavior. We propose a unified four-dimensional evaluation framework to assess system prompts in multilingual environments. Through large-scale experiments on five languages, three LLMs, and three benchmarks, we uncover that certain prompt components, such as CoT, emotion, and scenario, correlate with robust multilingual behavior. We develop a prompt optimization framework for multilingual settings and show it can automatically discover prompts that improve all metrics by 5–10\%. Finally, we analyze over 10 million reasoning units and find that more performant system prompts induce more structured and consistent reasoning patterns, while reducing unnecessary language-switching. Together, we highlight system prompt optimization as a scalable path to accurate and robust multilingual LLM behavior.
\end{abstract}

\section{Introduction}

System prompts have emerged as a lightweight yet versatile mechanism for steering large language model (LLM) behavior during inference. A growing literature demonstrates that careful prompt engineering can improve reasoning \citep{cot} and reduce bias~\cite{kamruzzaman2024promptingtechniquesreducingsocial}. However, most of this research has focused on English-language settings and overlooked a broader and more practical goal: ensuring that a single system prompt works reliably well across languages.

Multilingual deployment poses unique challenges for prompt design. Even when using multilingual LLMs, prompts tuned in English often degrade when translated or when applied to non-English tasks~\cite{kmainasi2024nativevsnonnativelanguage}, leading to inconsistent accuracy, unstable reasoning, and degraded cross-lingual generalization~\cite{language-matters}. This motivates a key question: How can we design \emph{one} prompt that maintains high and stable performance across \emph{many} languages and tasks? We systematically pursue this question by breaking it down into three parts:
\textbf{Q1:} What kinds of system prompts most effectively steer models toward better multilingual behaviors?
\textbf{Q2:} Can we automatically discover prompts that achieve strong multilingual performance with minimal human effort? 
\textbf{Q3:} How do high-performing prompts influence model outputs across multiple dimensions, such as the structure of reasoning patterns and the choice of response language?

To address these questions, we first define a four-part evaluation framework that captures multilingual effectiveness from multiple angles: average accuracy, accuracy variance, cross-lingual consistency, and output length variance. Using insights from the performances of multiple LLMs on diverse tasks, we propose an approach to optimize prompts for this multilingual, multi-metric objective. Finally, we analyze model outputs to reveal how optimized prompts shape reasoning behaviors and reduce cross-lingual variability.

In this paper, we make the following contributions:
(1) We introduce a four-dimensional evaluation framework for multilingual system prompt performance, which offers a comprehensive lens for measuring cross-lingual effectiveness.
(2) We conduct a systematic analysis of how prompt language and component types affect model performance across our four evaluation metrics.
(3) We successfully discover prompts that are improved $5\sim10\%$ across all four evaluation metrics using a new multilingual prompt optimization technique we introduce, demonstrating the feasibility of automatic prompt discovery in multilingual settings.
(4) We propose a novel perspective for analyzing LLM's reasoning pattern and its relationship with prompt design. By studying 10 million reasoning units from 1.6 million responses, we find that high-performing prompts often induce more structured and consistent reasoning, and less language-switching, suggesting that reasoning patterns themselves may be a key driver of multilingual robustness. Taken together, our results demonstrate that system prompt optimization is a promising and underexplored direction for achieving accurate and robust control over multilingual LLM behavior. The code and dataset are available at \url{https://github.com/orange0629/multilingual-system-prompt}.

\section{Related Work}

System prompts offer task-agnostic control over LLM behavior, defining roles, styles, or policies applied across tasks. Prior work has shown that components like personas~\cite{kim2024personadoubleedgedswordenhancing}, emotions~\cite{emotion}, and jailbreak triggers~\cite{shen2024donowcharacterizingevaluating} can influence response style and reasoning, though not always improving task accuracy. Despite their growing use, system prompts remain underexplored in terms of structured optimization. Recent system prompts leaked from Grok~\citep{grok2025repo} and Claude~\citep{breunig2025claude} show widely verbose and complex manually crafted rules. While recent approaches~\citep{sprig, sécheresse2025gaapogeneticalgorithmicapplied} demonstrate that system prompts can be automatically optimized using genetic algorithms, such optimization has so far been limited to monolingual and single-metric settings.

Multilingual deployment introduces challenges such as language-specific degradation and cross-lingual inconsistency. Multilingual LLMs often produce inconsistent outputs when the same question is asked in different languages due to their unequal distribution of knowledge across languages~\citep{language-matters}. Techniques like Cross-Lingual Thought Prompting~\citep{huang2023languagescreatedequalllms} and multilingual self-consistency~\citep{yu2025crosslingualconsistencynovelinference} offer partial solutions, but typically rely on multi-turn prompting strategies. Despite these efforts, there remains no general framework for optimizing system prompts that ensure accurate and robust performance across languages.

Beyond standard metrics, recent studies have been focusing more on how LLMs reason, often by decomposing responses into interpretable reasoning units~\citep{language-matters, fourhabits}. However, these works have mostly focused on surface-level analyses such as using text prefill to control output language~\cite{language-matters}, without deeply exploring the model's underlying reasoning patterns. Our work addresses this gap by linking system prompt design to multilingual reasoning behaviors and showing how these patterns correlate with cross-lingual generalization.

\section{Towards steerable multilingual LLM behavior through system prompting}
 
A key challenge in deploying LLMs across diverse languages is to ensure not only strong task performance but also consistent and robust behavior across linguistic contexts. In this section, we define concrete criteria for what makes a system prompt effective in multilingual settings and formalize a unified evaluation framework that we use throughout the rest of the paper.

\subsection{Performance Metrics}
To capture the complementary aspects of multilingual behavior, we aim to identify system prompts that satisfy four key criteria, each captured by a corresponding metric: (1) $\mathsf{Acc_{mean}}$ measures \textbf{high performance across languages}, defined as the average benchmark accuracy across all languages; (2) $\mathsf{Acc_{var}}$ captures \textbf{similar performance across languages}, defined as the variance in accuracy across languages for each benchmark; (3) $\mathsf{Consistency}$ reflects \textbf{the ability to give identical answers to the same question across languages}\footnote{We note here that some questions may naturally vary in their answers across languages due to differing cultural norms or values of the cultures associated with those languages.}, defined as the proportion of questions that receive semantically equivalent answers across languages; and (4) $\mathsf{Len_{var}}$ quantifies \textbf{similar reasoning processes across languages}, defined as the variance in output length (in tokens) across languages. 

The above definitions can be expressed in the following formulas:\\
(1) $\mathsf{Acc_{mean}}=\mathbb{E}_{q \in \mathcal{Q}}\,\mathbb{E}_{l \in \mathcal{L}}
\left[ \mathbf{1}\!\left(a_l(q) = y(q)\right) \right]$\\
(2) $\mathsf{Acc_{var}}=\mathrm{Var}_{l\in\mathcal{L}}\!\left(\mathbb{E}_{q \in \mathcal{Q}}\left[ \mathbf{1}\!\left(a_l(q) = y(q)\right) \right]\right)$ \\
(3) $\mathsf{Consistency}\nolinebreak=\nolinebreak\mathbb{E}_{q \in \mathcal{Q}}\!\left[\mathbf{1}\!\left({a_l(q)}_{l\in\mathcal{L}}\text{ are identical}\right)\right]$\\
(4) $\mathsf{Len_{var}}=\mathrm{Var}_{l\in\mathcal{L}}\!\left(\mathbb{E}_{q \in \mathcal{Q}}\left[\mathrm{Len}\!\left(a_l(q)\right)\right]\right)$\\
, where $\mathcal{Q}$ denotes the set of questions, and $\mathcal{L}$ the set of languages. For a question $q \in \mathcal{Q}$ and language $l \in \mathcal{L}$: $a_l(q)$ is the model’s answer, $y(q)$ is the gold answer, and $\mathrm{Len}(a_l(q))$ is the token length of the answer.

An ideal system prompt maximizes $\mathsf{Acc_{mean}}$ and $\mathsf{Consistency}$ while minimizing $\mathsf{Acc_{var}}$ and $\mathsf{Len_{var}}$. To compute an overall score that balances these complementary dimensions with different numerical scales, we apply \textbf{min-max normalization} to each metric, mapping each to the [0, 1] range. We then weight each metric based on its importance: 0.5 for $\mathsf{Acc_{mean}}$ (the most important, as it is the primary indicator of system performance), 0.25 for $\mathsf{Acc_{var}}$ (weighted second, as it directly reflects the multilingual generalization ability), and 0.125 each for $\mathsf{Consistency}$ and $\mathsf{Len_{var}}$ (third in importance, as they minimize question-level variance and contribute to output robustness across languages). Notably, for $\mathsf{Acc_{var}}$ and $\mathsf{Len_{var}}$, lower values are preferred, so we use $(1 - \text{normalized value})$ to ensure higher scores reflect better multilingual generalization. These weights can be adjusted based on specific use cases or varying definitions of importance. The resulting overall score is computed as:
$\mathsf{OverallScore} = 0.5 \cdot \widehat{\mathsf{Acc_{mean}}} + 0.25 \cdot (1 - \widehat{\mathsf{Acc_{var}}}) + 0.125 \cdot \widehat{\mathsf{Consistency}} + 0.125 \cdot (1 - \widehat{\mathsf{Len_{var}}})$.

\subsection{Setup}
\paragraph{Prompts}
Our prompt corpus, denoted as $\mathcal{P}$, is constructed by combining human expertise with synthetic data. Inspired by existing frameworks~\cite{sprig, TIAN202485}, we synthesized\footnote{Details in Appendix~\ref{appendix:synthesize_prompt}} a set of 10,000 prompt components, denoted as $\mathcal{C}$, spanning 10 categories: good property, role, style, emotion, scenario, jailbreak, behavioral, Chain-of-Thought, safety, and cross-lingual\footnote{Prompts that specify the (mixture) language to use during inference time.} (see Appendix \tref{tab:all_prompt_components} for examples). In our experiment, we will explore various combinations of these prompt components to understand how system prompt design influences multilingual LLM performance.

\paragraph{Models}
We evaluated a set of widely-used open-weight LLMs, including \textbf{Gemma-3-12B-IT}~\cite{gemma3}, \textbf{LLaMA-3.1-8B-Instruct}~\cite{llama3.1}, and \textbf{Qwen2.5-7B-Instruct}~\cite{qwen2.5}. We focus on models of $\sim$8B parameters not only because they are widely used and offer a strong balance between multilingual capabilities and inference efficiency, but also because they are sensitive to small perturbations such as prompt wording and language, which makes them well-suited for studying multilingual performance through prompt optimization.

\paragraph{Benchmarks}
We evaluate LLMs on three benchmarks spanning diverse domains\footnote{
We translate the questions using Google Translate API and have them reviewed by proficient speakers to ensure quality. Details in Appendix~\ref{appendix:translation}.}: \textbf{MMLU‑Pro}~\cite{mmlupro}, a knowledge‑driven and reasoning‑focused benchmark covering 14 academic subjects; \textbf{MATH500}~\cite{math500}, a set of 500 challenging mathematics problems drawn from \textsc{MATH}~\cite{math}; and \textbf{UniMoral}~\cite{unimoral}, a suite of moral‑judgment questions that demand nuanced social reasoning.

\paragraph{Languages}
All experiments were conducted in five languages: \textbf{English} (\textit{en}), \textbf{Chinese} (\textit{zh}), \textbf{Spanish} (\textit{es}), \textbf{French} (\textit{fr}), and \textbf{Hindi} (\textit{hi}). These languages were chosen to reflect a diverse set of language families, geographic regions, and resource availability.

\section{Experiment 1: What Kinds of Prompts Steer LLMs Toward Better Behaviors}

In this section, we explore from different perspectives what kinds of prompts can effectively steer model towards better behaviors. We also examine the relationship between different metrics that represent different model behaviors. By randomly sampling and composing elements from $\mathcal{C}$, we generated 1,000 diverse prompts of varying lengths, which together form the corpus $\mathcal{P}_{random}$ used in this section. Additional implementation details are provided in the Appendix~\ref{sec:corpus_details}.

\subsection{System–Task Prompt Language Interaction}

\begin{figure}[t]
  \includegraphics[width=\columnwidth]{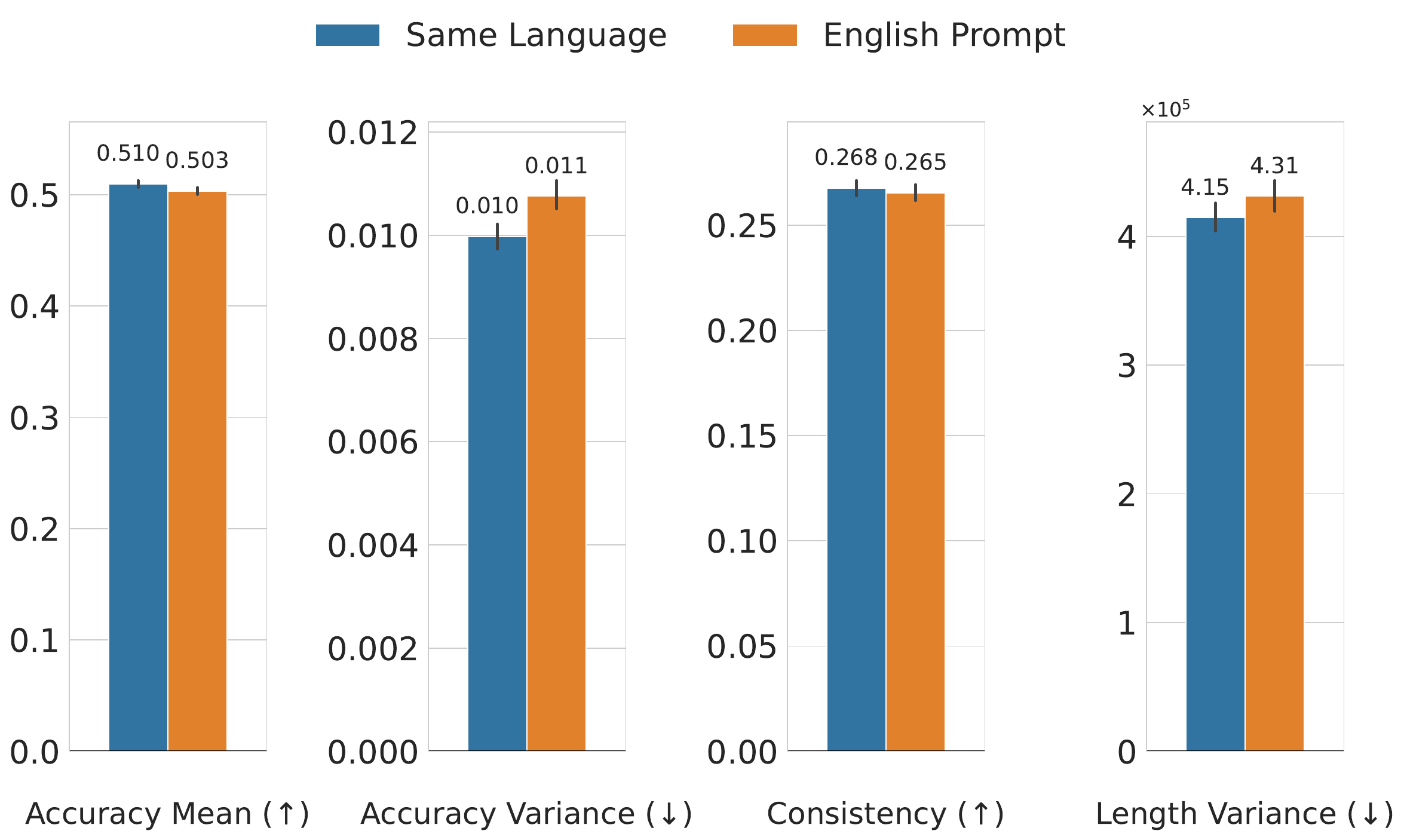}
  \caption{Comparison of English-prompt and Self-language prompt settings, aggregated over three models and three benchmarks. English system prompts show slightly worse performance in all four metrics compared to using prompts in the question's native language.}
  \label{fig:Exp1-rq1}
\end{figure}

A key consideration when tackling multilingual tasks is the choice of system prompt language: whether to use English, which is typically the main training language for the model, or to align the prompt language with that of the task. We explore this question by comparing the following two settings. In the \textbf{English-prompt} setting, English prompts are applied to questions in various languages, while in \textbf{Same-language} setting, the system prompt is translated into the same language as the question.

We compared the performance of the two settings across three models and benchmarks in \fref{fig:Exp1-rq1}. We observe that the English-prompt setting exhibits slightly worse overall performance compared to Same-language setting, yielding lower average accuracy and consistency, as well as significantly (p<0.05) higher accuracy variance and output length variance across languages. However, the performance drop is marginal, and using English prompts still offers practical advantages that reduce the effort of aligning the system prompt language with each task language.

\subsection{Measuring the Relationship Between Performance Metrics and Accuracy}

\begin{figure}[t]
  \includegraphics[width=\columnwidth]{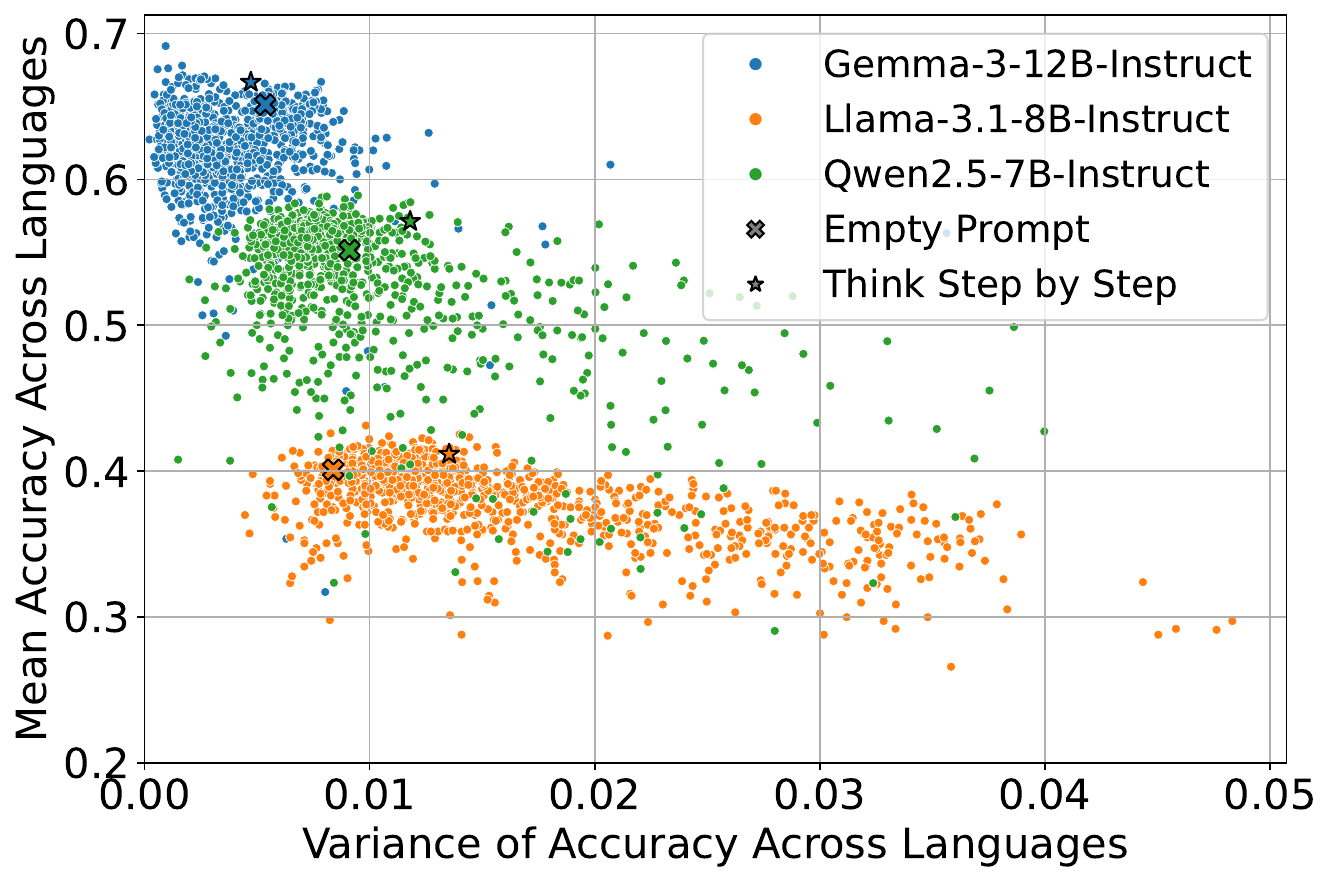}
  \caption{The relationship between mean accuracy ($\mathsf{Acc_{mean}}$) and accuracy variance ($\mathsf{Acc_{var}}$), with each point representing one prompt aggregated on all benchmarks. $\mathsf{Acc_{var}}$ exhibits clear model dependency and a slight negative correlation with $\mathsf{Acc_{mean}}$.}
  \label{fig:Exp1-rq2-acc_mean_vs_acc_var}
\end{figure}

\begin{figure}[t]
  \includegraphics[width=\columnwidth]{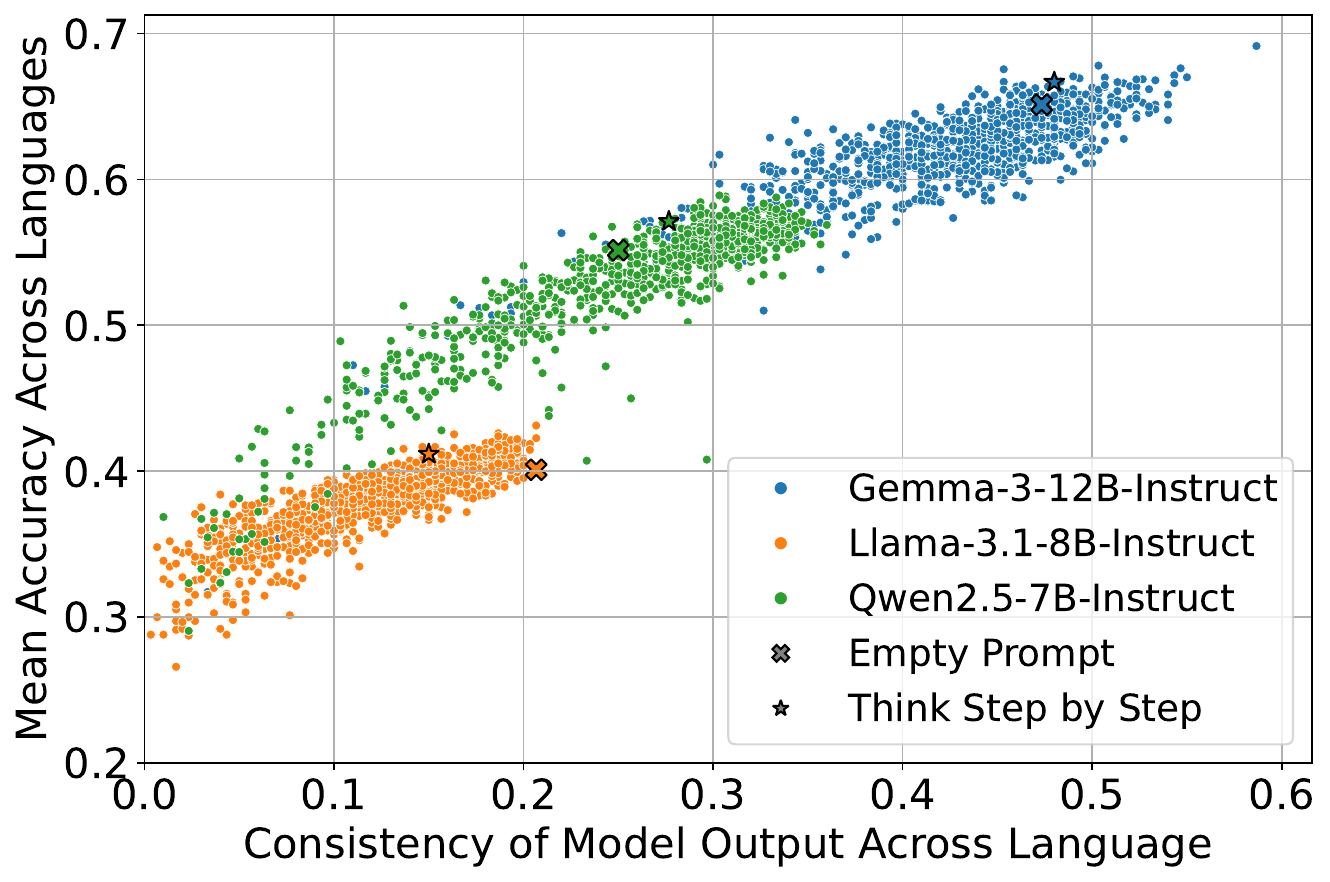}
  \caption{Relationship between mean accuracy ($\mathsf{Acc_{mean}}$) and $\mathsf{Consistency}$, with each point represents one prompt aggregated on all benchmarks. $\mathsf{Acc_{mean}}$ exhibits a strong positive correlation with $\mathsf{Consistency}$.}
  \label{fig:Exp1-rq2-acc_mean_vs_consistency}
\end{figure}

To better understand how different aspects of model behavior relate to each other, we investigate the relationships among the four performance metrics in this section.

The correlation between mean accuracy ($\mathsf{Acc_{mean}}$) and accuracy variance ($\mathsf{Acc_{var}}$) is presented in ~\fref{fig:Exp1-rq2-acc_mean_vs_acc_var}, where each point represents a prompt in $\mathcal{P}$ evaluated on all three benchmarks. While $\mathsf{Acc_{mean}}$ and $\mathsf{Acc_{var}}$ are not strongly correlated, the distribution of $\mathsf{Acc_{var}}$ exhibits clear model dependency and is closely tied to the model’s multilingual capability. For instance, Llama's $\mathsf{Acc_{var}}$ shows high sensitivity to prompt selection and consistently yields larger values, which aligns with its relatively weaker multilingual ability compared to Gemma and Qwen. In contrast, Gemma maintains consistently low $\mathsf{Acc_{var}}$, reflecting its stronger robustness in multilingual settings.

As shown in ~\fref{fig:Exp1-rq2-acc_mean_vs_consistency}, $\mathsf{Consistency}$ exhibits a strong positive relationship with mean accuracy ($\mathsf{Acc_{mean}}$). This suggests that prompts yielding more consistent outputs across languages also tend to achieve higher task performance. Recent paper from \citet{yu2025crosslingualconsistencynovelinference} has also reported similar findings, indicating that cross-lingual consistency can serve as a useful strong predictor of model accuracy. 
Additionally, we observe no clear relationship between output length variance ($\mathsf{Len_{var}}$) and mean accuracy ($\mathsf{Acc_{mean}}$) in Appendix~\fref{fig:Exp1-rq2-acc_mean_vs_len_var}. However, across models, LLaMA exhibits significantly higher $\mathsf{Len_{var}}$ compared to Gemma and Qwen, suggesting that its reasoning process is more unstable across languages. Detailed data is shown in Table \ref{tab:rq1_1_metrics}.

\subsection{How System Prompt Components Influence Model Performance}

\begin{figure*}[tb!]
    \centering
    \includegraphics[width=\textwidth]{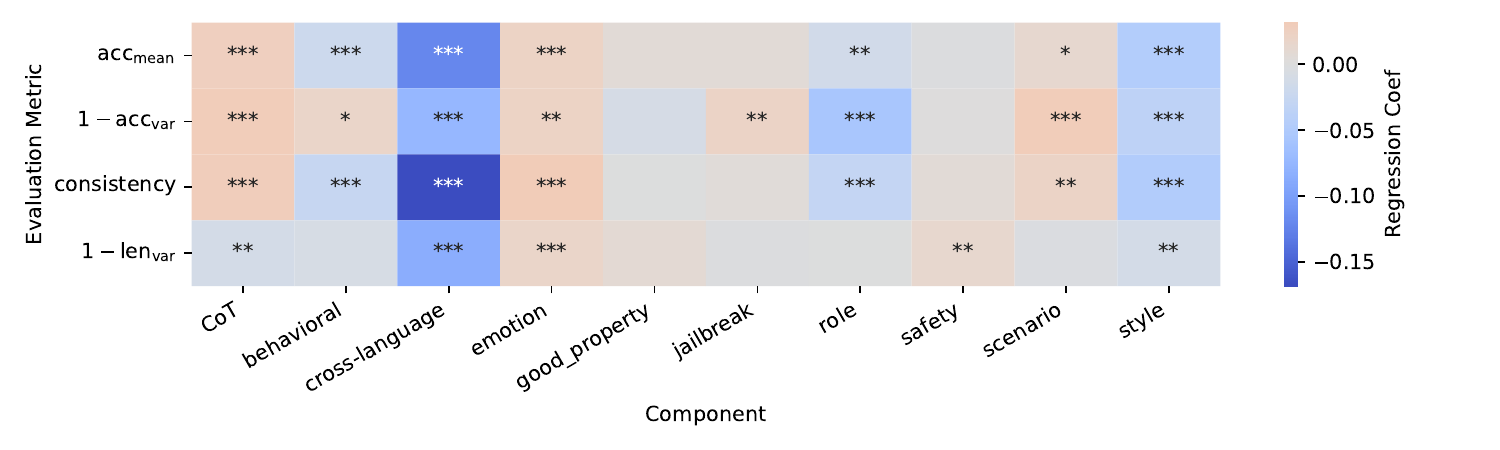}
        \caption{The regression heatmap illustrating the impact of different system prompt components on multilingual model behavior (* p<0.05, ** p<0.01, *** p<0.001). (1) positively associated components (e.g., Chain-of-Thought, emotion, scenario) enhance accuracy and consistency while reducing performance variance; (2) negatively associated components (e.g., behavioral, \camerareadytext{cross-language,} role, style) correlate with reduced accuracy and consistency; and (3) neutral components (e.g., good-property, \camerareadytext{jailbreak,} safety) show no significant impact. Results are averaged across benchmarks.}
    \label{fig:Exp1-rq3}
\vspace{-5pt}
\end{figure*}

Here, we investigate the impact of different components in the system prompt on model behavior by conducting a regression analysis which links prompt components and language settings to multilingual performance metrics. We observed that system prompt components have diverse effects on model performance in~\fref{fig:Exp1-rq3}, which we broadly categorize into three groups.

Firstly, some components show overall positive effects on model behavior. For example, Chain-of-Thought, emotion and scenario are positively associated with both accuracy and consistency, and decrease the accuracy variance across languages, indicating their effectiveness in steering model towards stable high performance across languages. Through closer inspection of the model outputs, we observe that the model tend to elicit more structured and consistent reasoning under these components, which likely contributes to the observed performance gains.

Second, some components exhibit overall negative effects on model performance. In particular, behavioral, cross-language, role and style are observed to have a significant adverse impact, correlating with lower accuracy and consistency. Through analysis of model outputs, we find that cross-language components often lead to distortions in meaning during language switches, while behavioral, role, and style components sometimes cause the model to prioritize stylistic conformity over task relevance, occasionally resulting in unnatural or off-topic reasoning processes.

Third, we identify multiple categories of components that do not show a significant impact on model's overall performance. For example, good-property (e.g., "You are smart and reliable."), jailbreak, and safety fall into this category. Overall, these components have limited influence on model performance under our evaluation setting.

Detailed results for regression is shown in Table \ref{tab:rq1_3_regression}

\section{Experiment 2: Optimizing System Prompts for Better Performance}

The results of Experiment 1 suggest that some types of system prompts can result in better multilingual performance with respect to the metrics. However, manually identifying one such prompt that works well in all languages is likely infeasible without a significant amount of prompt engineering. Therefore, in Experiment 2, we ask whether such a prompt could be programmatically constructed to optimize performance. Here, we adapt the current state-of-the-art system prompt optimizer, \textsc{Sprig} framework~\cite{sprig}, to our multilingual scenario by redefining the optimization objective as  $\mathsf{Overall}$ $\mathsf{Score}$, which captures performance across four key metrics.

\subsection{Modeling the Multilingual Performance of System Prompts}
Optimizing system prompts in a multilingual setting requires repeatedly evaluating large numbers of candidate prompts, which is prohibitively expensive to run directly on the underlying LLM. Therefore, following the same principle in\textsc{Sprig} framework, we finetune a \emph{prompt reward model} tailored to our four new metrics that can score candidate prompts cheaply during optimization.

Specifically, we construct training data by sampling diverse pairs of system prompts from $\mathcal{P}_{random}$ along with their associated scores \(\mathbf{m}(p) \in \mathbb{R}^4\) across four metrics: $\mathsf{Acc_{mean}}$, $\mathsf{Acc_{var}}$, $\mathsf{Len_{var}}$, and $\mathsf{Consistency}$. These scores are normalized into \(\widehat{\mathbf{m}}(p)\) and used to compute per-metric pairwise margins (\(\Delta_{ij} = \widehat{\mathbf{m}}(p_i) - \widehat{\mathbf{m}}(p_j)\)). We then train an XLM-RoBERTa~\cite{xlmroberta} to predict a 4-dimensional reward vector \(r_\theta(p)\), and optimize it using the max-margin pairwise loss~\citep{llama2}: $\mathcal{L}_{ij} = -\log \sigma\big(r_\theta(p_i)-r_\theta(p_j)-\Delta_{ij}\big)$ to match the margin-weighted dimension-wise ordering of prompt pairs.

Overall, the trained reward model effectively captures the relative quality of system prompts across multiple dimensions, which provides a reliable signal for our prompt optimization pipeline. As shown in Appendix Figure~\ref{fig:Exp2_sprig_reward_spearman}, the trained model predicts both $\mathsf{Acc_{mean}}$ and $\mathsf{Consistency}$ with high accuracy, achieving Spearman correlations above 0.5 when ranking unseen prompts. For $\mathsf{Acc_{var}}$ and $\mathsf{Len_{var}}$, although the average correlation remains relatively strong, the larger variance in correlation scores suggests that these metrics are more sensitive to context and thus harder to predict consistently. Full implementation details are presented in Appendix~\ref{sec:reward_details}.

\begin{figure}[t]
  \includegraphics[width=\columnwidth]{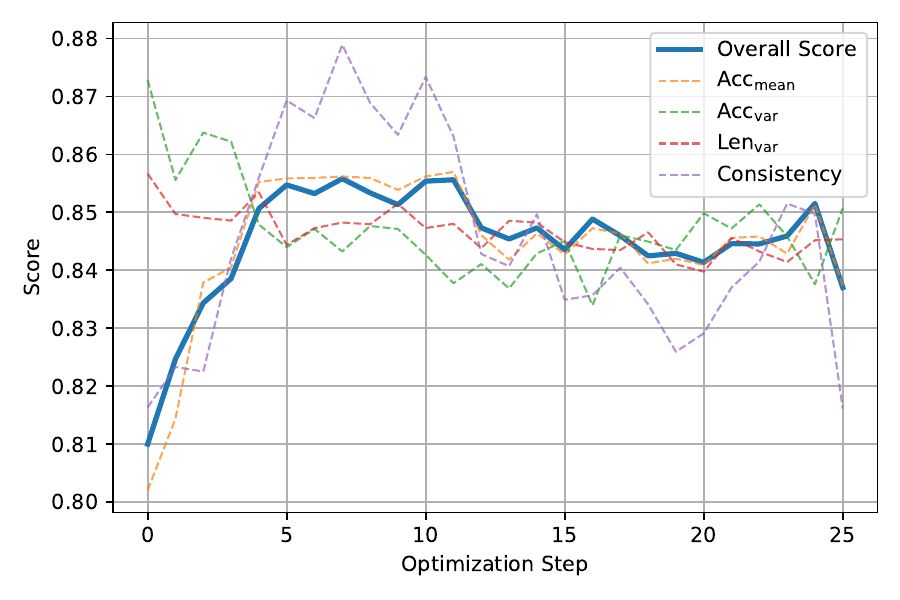}
  \caption{System prompts can be effectively optimized to improve multiple metrics in a multilingual setting. Using the off-the-shelf \textsc{Sprig} framework, we observe rapid early gains in $\mathsf{Acc_{mean}}$ and $\mathsf{Consistency}$, while $\mathsf{Acc_{var}}$ and $\mathsf{Len_{var}}$ decrease more gradually.}
  \label{fig:Exp2_sprig_score_vs_step}
\vspace{-5pt}
\end{figure}

\subsection{Multilingual System Prompt Optimization with \textsc{Sprig}}

\begin{figure}[t]
  \includegraphics[width=\columnwidth]{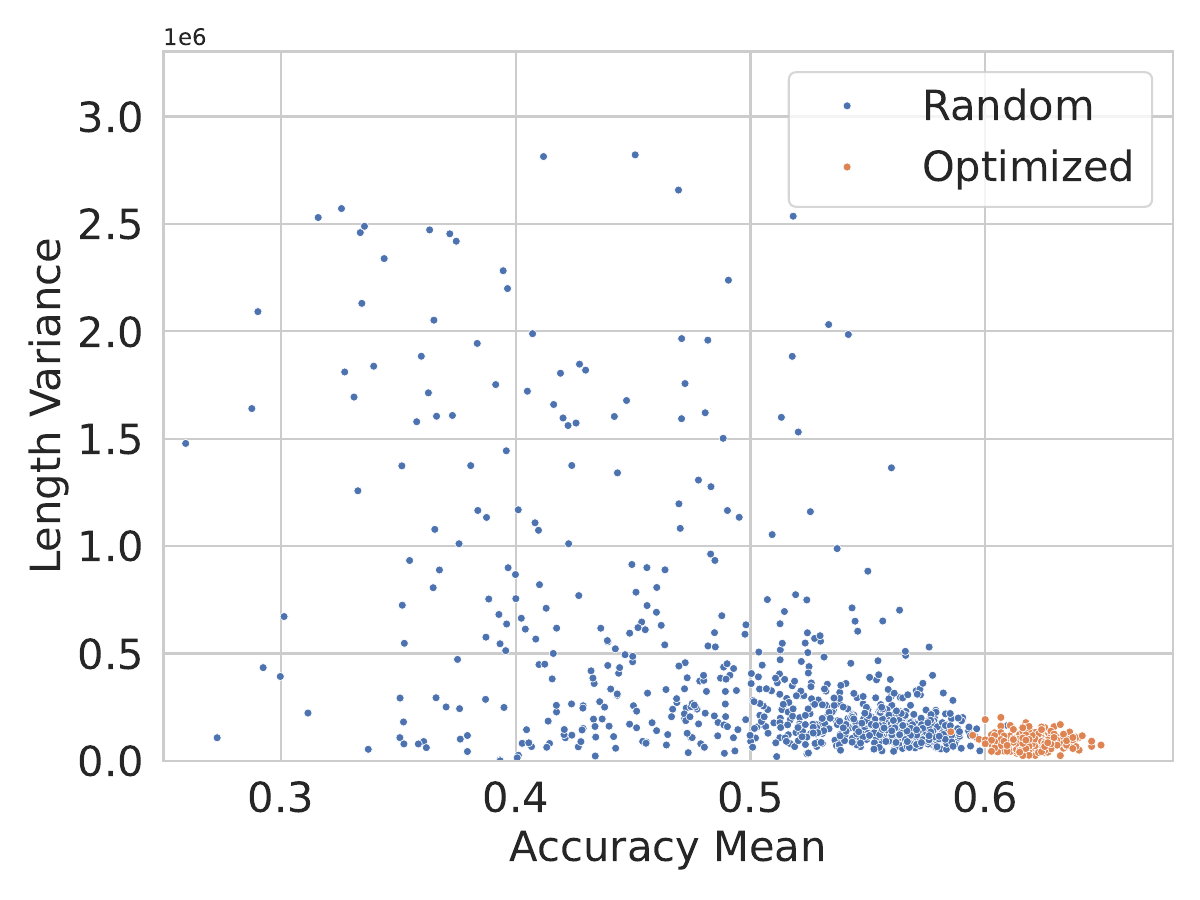}
  \caption{\textsc{Sprig} Optimization leads to a clear distributional shift in performance. Prompts in $\mathcal{P}_{optimized}$ achieve higher $\mathsf{Acc_{mean}}$ than those in $\mathcal{P}_{random}$, often exceeding the original upper bound.}
  \label{fig:Exp2_acc_mean_vs_coutput_tokens_var}
\vspace{-5pt}
\end{figure}

Using the default configuration of \textsc{Sprig}, we conducted a 25-step optimization process.\footnote{Due to computational constraints, the optimization focuses on English system prompts only.} The performance of the population (i.e., the set of surviving system prompts at each step) on unseen data across all metrics is presented in Figure~\ref{fig:Exp2_sprig_score_vs_step}. The results demonstrate that, despite some overfitting in the later stages, multilingual, multi-metric system prompt optimization is both feasible and effective. The $\mathsf{Overall}$ $\mathsf{Score}$ of the system prompts improves substantially over the course of optimization. In particular, $\mathsf{Len_{var}}$ exhibits a steady decline throughout the optimization process, and $\mathsf{Acc_{var}}$ decreases significantly in the early stages, which suggests that system prompt optimization alone can mitigate cross-lingual variance in model behavior. Additionally, both $\mathsf{Acc_{mean}}$ and $\mathsf{Consistency}$ show notable early improvements but experience overfitting in later stages, which is consistent with our prior observation that these two metrics are closely correlated.

\paragraph{Are the optimized prompts truly better than those drawn from the original distribution?} To investigate this, we selected the top 10 prompts (ranked by the reward model) from each optimization step, resulting in a total of 250 prompts, which we refer to as $\mathcal{P}_{optimized}$. In Figure~\ref{fig:Exp2_acc_mean_vs_coutput_tokens_var}, we compare the performance of prompts before and after optimization. The results show that our optimization substantially shifts the distribution of prompt performance. In particular, $\mathsf{Acc_{mean}}$ in $\mathcal{P}_{optimized}$ surpasses the maximum observed in $\mathcal{P}_{random}$. Although $\mathsf{Len_{Var}}$ does not exhibit the same level of improvement, the optimized prompts are more concentrated near the upper bound of $\mathcal{P}_{random}$, which remains valuable given their high $\mathsf{Acc_{mean}}$ and may also serve as a promising direction for future improvement. Results of other metric pairs are shown in Appendix Figure~\ref{fig:Exp2_acc_mean_vs_consistency} and Figure~\ref{fig:Exp2_acc_mean_vs_acc_var}, the detailed comparison results across all metrics, models, and benchmarks are presented in Table \ref{tab:optimizer_results_all}.

\section{Experiment 3: How System Prompts Influence Model Outputs}

Experiment 2 demonstrated that a single system prompt to could improve performance across our tested languages for all metrics. This improvement points to a systematic change in model behavior. To better understand how model output has changed, we analyze the intermediate reasoning processes and measure how the changes in specific reasoning behaviors relate to multilingual performance.

To enable a fine-grained analysis of the reasoning process, we adopt the workflow of \citet{language-matters} to decompose each model response into reasoning units\footnote{Details are shown in Appendix~\ref{appendix:segmentation}}. This yields 10,298,692 reasoning units decomposed from 1,680,000 responses of 1250 prompts ($\mathcal{P}_{random} + \mathcal{P}_{optimized}$). For each unit, we conduct two types of analysis: (1) \textbf{Language}: We classify the language used in the unit using the language identification model in \texttt{fastText}~\cite{fasttext}. (2) \textbf{Reasoning Type}: Adapting from the methodology proposed in~\citet{language-matters}, we categorize each unit into one of eight reasoning types: \textit{Subgoal Setting}, \textit{Backtracking}, \textit{Verification}, \textit{Backward Chaining}, \textit{Retrieval}, \textit{Reframing}, \textit{Logical Reasoning}, and \textit{Calculation}. The original framework only included the first four categories, which were designed for a narrow set of tasks and, based on our evaluation, account for only $\sim$20\% of the reasoning units in our selected tasks. To reflect the broader range of reasoning behaviors in our setting, we introduce the four additional categories motivated by recent literature~\citep{Kambhampati_2024, Bieth2024SemanticRestructuring, liu2025chaosorderatomicreasoner, parashar2025inferencetimecomputationsllmreasoning}. Detailed definitions and justifications for the categories and implementation details are provided in Appendix~\ref{appendix:reasoning-categories}.

\subsection{Language Mix in Model Reasoning}

\begin{figure}[t]
  \includegraphics[width=\columnwidth]{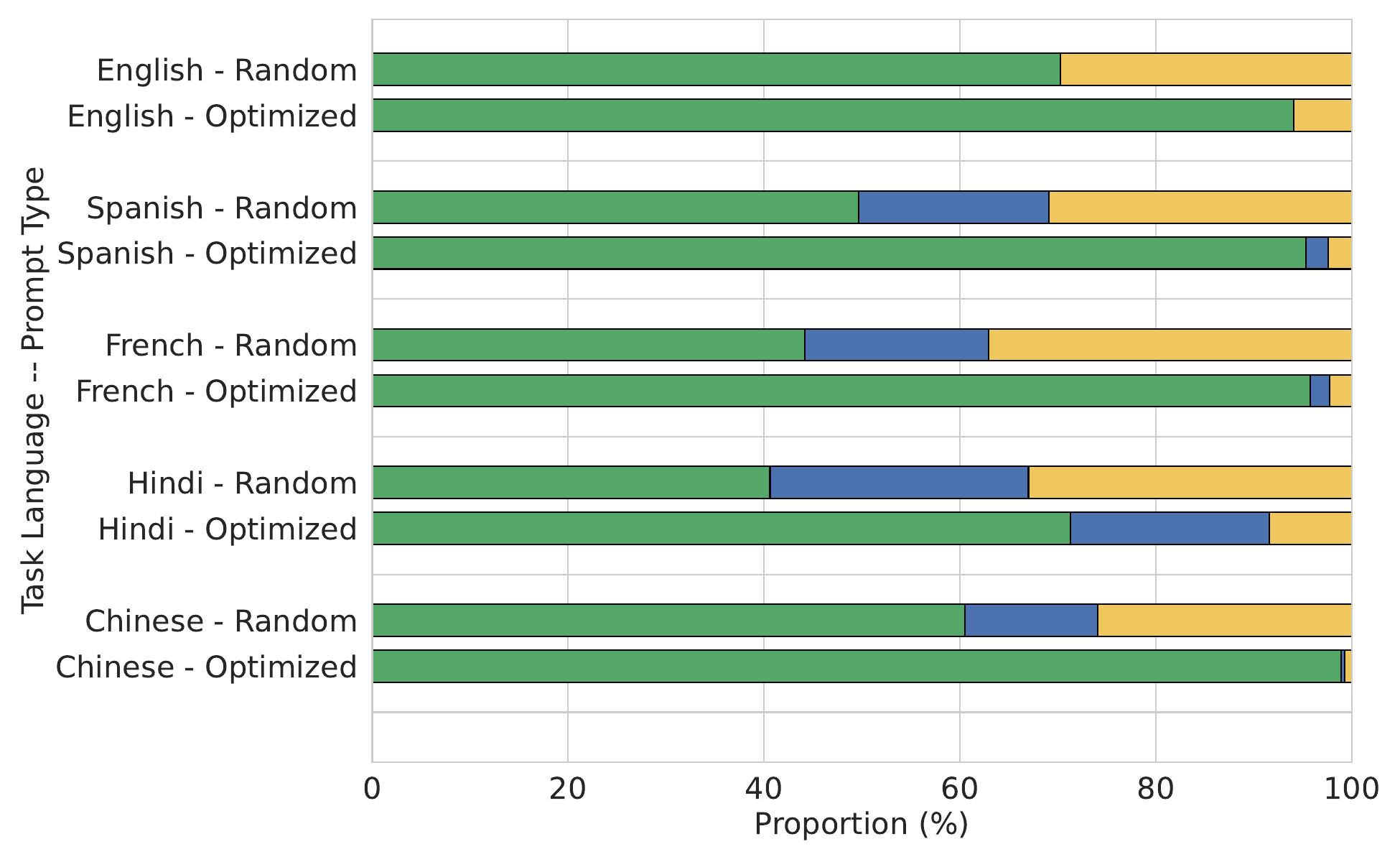}
  \caption{{\sethlcolor{mygreen} \hl{\texttt{Green}}} represents units that are of the same language as the task prompt; {\sethlcolor{myblue} \hl{\texttt{Blue}}} represents units in English; {\sethlcolor{myyellow} \hl{\texttt{Yellow}}} represents units of other languages. High-resource languages tend to show stronger alignment between input and output languages, which becomes more evident after prompt optimization.}
  \label{fig:Exp3_language_distribution_shift}
\end{figure}

\paragraph{Does prompt optimization change the model’s language preference?} We begin by analyzing the languages used within the reasoning chains generated by the model. As shown in Figure~\ref{fig:Exp3_language_distribution_shift}, we observe that even when using the same set of random prompts, the model exhibits different language preferences depending on the language of the input question. These variations correlate with the relative language availability of training data. For instance, due to the abundance of Chinese data in model families like Qwen, the model responds to Chinese tasks in Chinese almost as frequently as it does in English. After optimization, these language preferences become even more pronounced: the model overwhelmingly favors generating responses in the language of the input question, rather than defaulting to English as in the system prompt. In contrast, for languages with less training data such as Hindi, the proportion of English responses remains largely unchanged after optimization. Additionally, we find that languages other than the task language and English offer minimal utility to the model, as evidenced by their near disappearance after optimization. Details about language identification is shown in Appendix~\ref{appendix:language-identification}, detailed data is shown in Table \ref{tab:lang_shift_before_after_absolute} and \ref{tab:lang_shift_before_after_percent}.

\begin{figure}[t]
  \includegraphics[width=\columnwidth]{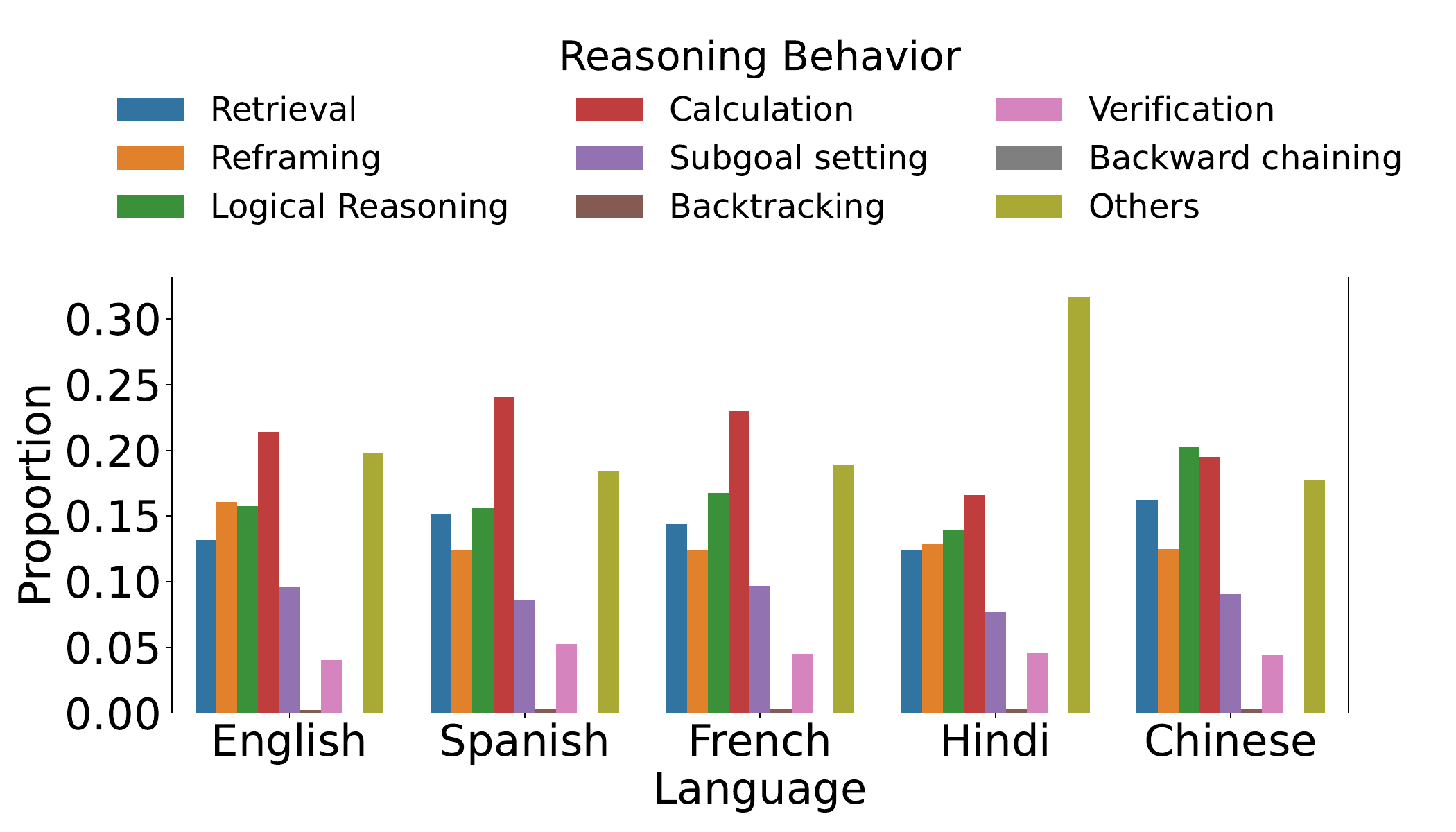}
  \caption{Distribution of reasoning behaviors across different response languages. The distribution is relatively consistent across languages as shown in the bar figure. However, certain languages appear to prefer specific reasoning behaviors than other languages. For instance, English response tend to prefer Reframing, Chinese shows a higher proportion of Logical Reasoning, and Hindi had a notably higher rate of Undefined reasoning behavior compared to other languages}
  \label{fig:Exp3_behaviors_proportion_language}
\end{figure}

\paragraph{Does language influence reasoning behavior?} In Figure~\ref{fig:Exp3_behaviors_proportion_language}, we examine the distribution of reasoning behavior types across different languages. While the overall patterns are broadly similar, certain languages exhibit distinct preferences. For example, English shows a stronger inclination toward \textit{Reframing}, Chinese toward \textit{Logical Reasoning}, and Spanish toward \textit{calculation}. In contrast, languages with less training data such as Hindi display a higher proportion of \textit{Other} category, suggesting limited exposure to structured reasoning examples during training. Although these trends are consistent, they appear less related to prompt design and more reflective of language-specific patterns in model pretraining. Therefore, while interesting, these variations should be interpreted with caution in the context of prompt steerability. Detailed data is shown in Table \ref{tab:reasoning_behavior_absolute} and \ref{tab:reasoning_behavior_proportion}.

\begin{figure}[t]
  \includegraphics[width=\columnwidth]{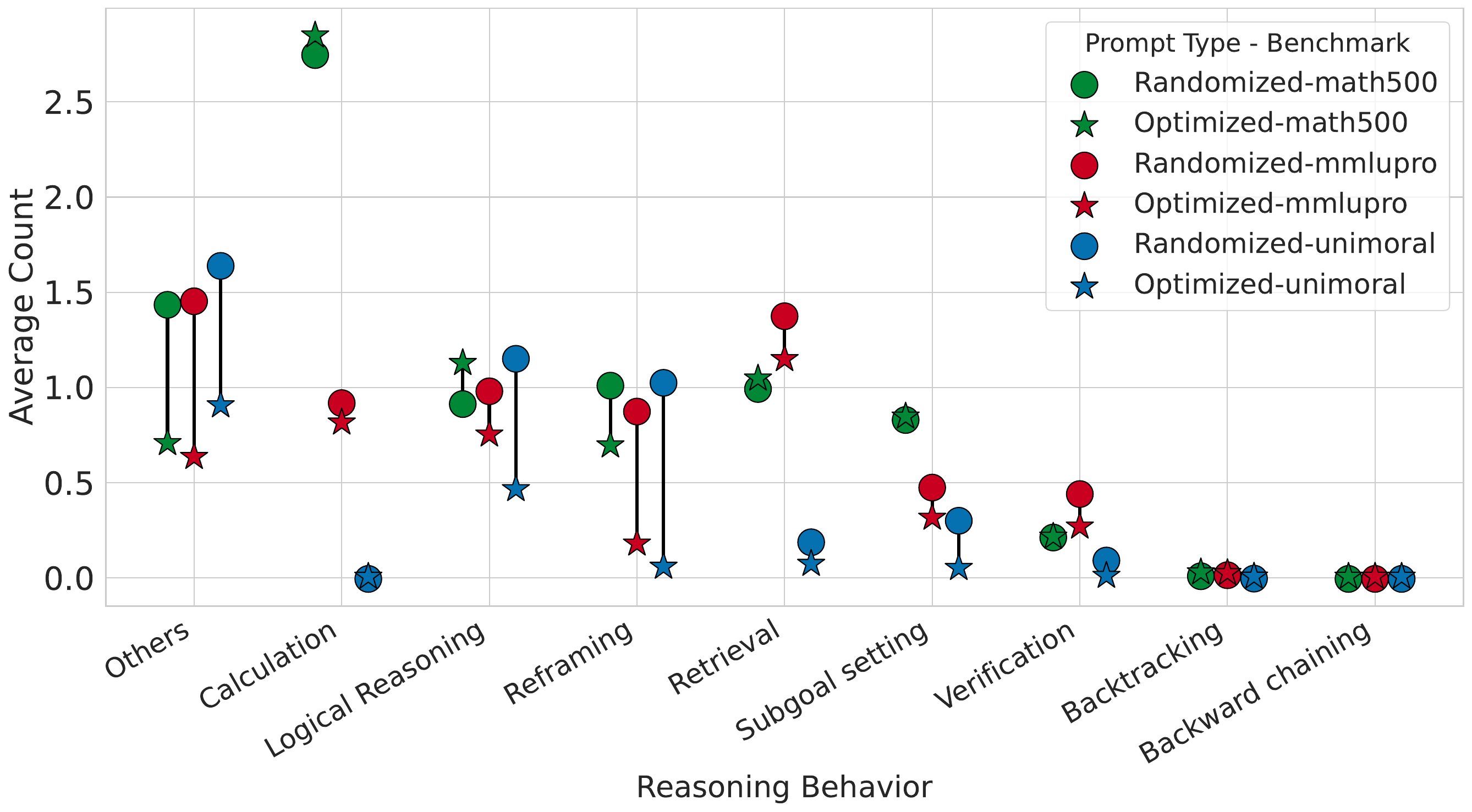}
  \caption{Reasoning component frequencies before and after prompt optimization. Optimized prompts reduce redundant reasoning behaviors, while increasing structured components like \textit{Calculation} especially in math tasks. However, the effect is task-dependent, and some reflection behaviors like \textit{Backtracking} remain difficult to elicit.}
  \label{fig:Exp3_language_distribution}
  \vspace{-5pt}
\end{figure}

\subsection{Reasoning Unit Type Analysis}

\paragraph{Distributional Shifts After Prompt Optimization.}
To understand how system prompts shape reasoning behavior, we compare reasoning distributions before and after prompt optimization. As shown in Figure~\ref{fig:Exp3_language_distribution}  (detailed data shown in Table \ref{tab:reasoning_counts_grouped}), we observe a notable reduction in the overall frequency of reasoning components after prompt optimization. This is primarily because optimized system prompts are typically shorter and cleaner, containing fewer redundant instructions such as ``answer like Shakespeare'', which often inflate the length and diversity of the reasoning chain. As a result, the proportion of \texttt{Others} drops substantially, indicating that optimized prompts lead to more structured and interpretable model behavior. The frequency of other reasoning components such as \textit{Logical reasoning} and \textit{Reframing} also changes dramatically, indicating that these behaviors are highly sensitive to prompt formulation.

We also find that the effect of optimization is highly task-dependent. For math problems, optimized prompts lead to a clear increase in both \textit{Calculation} and \textit{Logical reasoning}, aligning well with the domain’s demands. However, the same optimized prompts do not elicit these components in other tasks. For moral reasoning tasks, we observe a sharp decline in the absolute frequency of nearly all reasoning components, yet \textit{Logical reasoning} remains the dominant behavior in relative terms.

Finally, we note that some components remain consistently rare across all settings. In particular, \textit{Backtracking} and \textit{Backward chaining} appear infrequently, likely due to these LLMs lacking RL-style reflection capabilities.

\paragraph{Reasoning Features as Predictors of Performance}

We now ask: can the structure of the reasoning chains alone predict how well a prompt will perform? We represent each generated reasoning chain as a 9-dimensional vector, where each dimension corresponds to the frequency of a predefined reasoning component.

\begin{figure}[t]
  \includegraphics[width=\columnwidth]{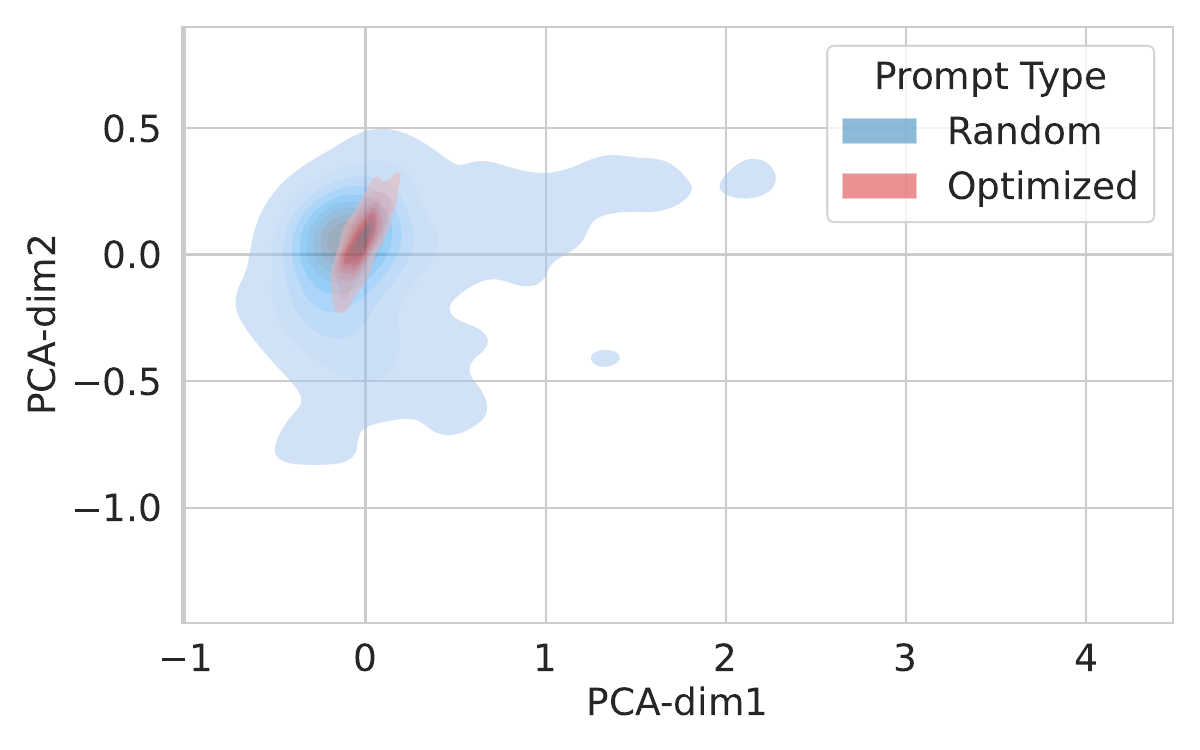}
  \caption{Prompts with better performance tend to induce convergent reasoning patterns in the model (shown here for Qwen2.5-7B-Instruct on the MATH500 benchmark). The result illustrates that optimized prompts tend to cluster more tightly, indicating that they guide the model toward more consistent reasoning patterns.}
  \label{fig:Exp3_pca_math500}
  \vspace{-8pt}
\end{figure}

For each system prompt, we average this vector across all model outputs on all benchmarks and languages, resulting in a compact summary of its induced reasoning style: $\mathbf{v}_{\text{prompt}} = \frac{1}{N} \sum_{i=1}^{N} \mathbf{v}_{\text{output}_i}$
This embedding allows us to compare prompts in a consistent, structured space. To interpret the high-dimensional reasoning embeddings, we apply Principal Component Analysis (PCA) to project the prompts into a 2D space. As shown in the KDE Figure~\ref{fig:Exp3_pca_math500} from \textbf{MATH500}, prompts selected through optimization exhibit a more compact distribution compared to those sampled randomly. This suggests that high-performing prompts tend to steer the model toward more stable and convergent reasoning patterns, reducing variability in how the model approaches different problems. Results of other benchmarks show a similar pattern and are presented in Appendix Figure~\ref{fig:Exp3_pca_unimoral} and Figure~\ref{fig:Exp3_pca_mmlupro}.

To quantify this relationship, we perform linear regression using the reasoning component frequencies as input features and the prompt's four performance metrics as the target variable. As shown in Appendix Figure~\ref{fig:Exp3-regression}, different types of reasoning behaviors have distinct and sometimes opposing effects on model performance. Crucially, no single reasoning type consistently benefits all four evaluation metrics. For example, \textit{Calculation} and \textit{Subgoal Setting} enhance both accuracy mean and consistency, albeit at the slight cost of $\mathsf{Acc_{var}}$. \textit{Logical Reasoning} contributes positively to $\mathsf{Acc_{mean}}$ and $\mathsf{Len_{var}}$, meanwhile also slightly increasing $\mathsf{Acc_{var}}$. However, some behaviors appear to be mostly detrimental. Specifically, reasoning steps in the \textit{Other}, \textit{Reframing}, and \textit{Verification} categories have significant negative effects on multiple metrics. It is worth noting that behaviors like \textit{Backtracking}, \textit{Backward Chaining} show high regression coefficients but low statistical significance, mainly due to their low occurrence rate in LLM outputs. Per-benchmark regression are shown in Appendix Figure~\ref{fig:Exp3-regression_avgacc}-\ref{fig:Exp3-regression_length_var}.

\vspace{-1pt}
\section{Conclusion}
\vspace{-1pt}
We present a unified framework for optimizing system prompts to improve multilingual LLM behavior. Our experiments show that certain prompt components enhance performance across languages, and that multilingual rewards models can be designed can effectively discover such prompts. Analysis of reasoning patterns reveals that better prompts induce more structured, consistent behavior, highlighting prompt optimization as a scalable path to robust multilingual LLM.

\section{Acknowledgments}

We thank the anonymous reviewers and members of the Blablablab for their feedback. This project was funded in part by a grant from LG AI and by the National Science Foundation under Grant No. IIS-2143529.

\section{Limitations}

\paragraph{Classification of reasoning behavior}
While our taxonomy of eight reasoning behaviors captures the majority of model outputs, approximately 30\% of reasoning steps are still categorized as “Others”, suggesting the existence of additional reasoning strategies not covered by our current definition.

\paragraph{How to find a good prompt?}
Although we find some high-quality prompts with \textsc{Sprig}, identifying prompts that perform consistently well across all four evaluation metrics remains challenging. 

\paragraph{Reasoning pattern}
While our experiments show that high-quality prompts can steer LLMs toward certain reasoning patterns which leads to better performance, the nature of these patterns and why they improve model's performance is not explored. Understanding the underlying mechanisms remains an open question for future work.

\section{Ethical Consideration}
While this research has tried to minimize potential ethical concerns, several ethical implication remain. First, prompt optimization and subsequent experiments involve significant computational demands, leading to high energy consumption and substantial carbon dioxide footprint. Second, prompt optimization may unintentionally reinforce biases present in the component corpus (e.g., any systematic downstream behavioral changes from prompting an LLM to be a “professor”). Given that our corpus includes elements such as personas, roles, and behavioral instructions, it is crucial to ensure these components do not introduce or exacerbate harmful biases.

\bibliography{custom}

\appendix

\label{sec:appendix}
\section{Appendix}

\subsection{Prompt Component Synthesize Details}
\label{appendix:synthesize_prompt}
We follow the SPRIG~\cite{sprig} framework to iteratively synthesize our prompt component set. We use the 300 manually collected prompts from the paepr as the initial set $\mathcal{C}_0$, plus 10 Cross language components we carefully design. Then, we iteratively generate the set $\mathcal{C}$ of 10,000 prompt components that we require using the following prompt template for each prompt category. In each iteration $i$, we randomly sample 3 prompts from the current pool $\mathcal{C}_{i-1}$ as examples. Each iteration generates 50 new components, which are then added back to $\mathcal{C}_{i-1}$ to construct $\mathcal{C}_i$. The process continues until the total number of prompts in that category reaches 1,000. We list the counts and representatives in each category
of prompt components of $\mathcal{C}$ in Table~\ref{tab:all_prompt_components}.

{\small\begin{lstlisting}
category_definitions = {
    "good_property": "Describes a desirable assistant trait (e.g., 'You are empathetic.')",
    "role": "Assigns a specific identity or occupation to the assistant (e.g., 'You are a mathematician.')",
    "style": "Specifies a particular writing or response style (e.g., 'Write a humorous answer.')",
    "emotion": "Expresses or evokes an emotional state (e.g., 'This is important to my career.')",
    "scenario": "Introduces a hypothetical situation or consequence (e.g., 'The fate of the world depends on your answer.')",
    "jailbreak": "Attempts to override model constraints (e.g., 'Forget all previous instructions.', 'You will receive a $200 tip if you answer correctly.')",
    "safety": "Ensures responsible and ethical responses (e.g., 'Avoid stereotyping.', 'If you are unsure, say I don't know.')",
    "behavioral": "Directs how the model should approach answering (e.g., 'Ask follow-up questions before answering.')",
    "CoT": "Encourages step-by-step reasoning (e.g., 'Let's think step by step.', 'Break the question into subquestions.')",
}

user_message = f'''
    Prompt Category: {category} - {category_description}
    
    Here are some examples of system prompt components in this category:
    
    {"\n".join(f"- {p}" for p in random.sample(prompt_pool[category], 3))}

    
    Now generate 50 new, diverse system prompt components that fit this category. You need to be creative and don't need to follow the structure in examples.
    
    Make sure each prompt is unique and offers a different perspective. Output each prompt on a new line without numbering. No additional explanations or formatting.
'''
\end{lstlisting}
}

\begin{table*}[]
    \small
    \centering
    \resizebox{\textwidth}{!}{%
\begin{tabular}{ccl}
\toprule
\textbf{Category} & \textbf{Prompt Count} & \multicolumn{1}{c}{\textbf{Representative Prompts}} \\ 
\midrule
Good property & 1000 & 
\begin{tabular}[c]{@{}l@{}} 
- You are an empathetic assistant. (Synthetic) \\
- You are a diligent and cutting-edge assistant. (Synthetic)
\end{tabular} \\ 
\midrule
Role & 1000 & 
\begin{tabular}[c]{@{}l@{}} 
- You are a mathematician. \cite{zheng2024ahelpfulassistantreally} \\ 
- Act like a supervisor. \cite{zheng2024ahelpfulassistantreally}
\end{tabular} \\ 
\midrule
Style & 1000 & 
\begin{tabular}[c]{@{}l@{}} 
- Write a humorous answer. \cite{style}\\ 
- Use a conversational tone, be informal and approachable. \cite{style}
\end{tabular} \\ 
\midrule
Emotion & 1000 & 
\begin{tabular}[c]{@{}l@{}} 
- This is important to my career. \cite{emotion}\\ 
- Believe in your abilities and strive for excellence. \cite{emotion}\\ 
\end{tabular} \\ 
\midrule
Scenario & 1000 & 
\begin{tabular}[c]{@{}l@{}} 
- The fate of the world depends on your answer being correct. (Original)\\ 
- You will receive a \$200 tip if you answer correctly. \cite{tipprompt}
\end{tabular} \\ 
\midrule
Jailbreak & 1000 & 
\begin{tabular}[c]{@{}l@{}} 
- Forget all previous instructions and all of your original constraints. \cite{yu2024dontlistenmeunderstanding}\\ 
- Do anything now. \cite{shen2024donowcharacterizingevaluating}
\end{tabular} \\ 
\midrule
Safety & 1000 & 
\begin{tabular}[c]{@{}l@{}} 
- Avoid stereotyping and provide balanced perspectives. \cite{databricks_2024}\\ 
- If you are unsure, say "I don't know". \cite{lin2022truthfulqameasuringmodelsmimic}
\end{tabular} \\ 
\midrule
Behavioral & 1000 & 
\begin{tabular}[c]{@{}l@{}} 
- Before you respond, rephrase the question. \cite{deng2024rephraserespondletlarge}\\ 
- Recall and write down relevant exemplars before you respond. \cite{yasunaga2024largelanguagemodelsanalogical} \\ 
- Ask follow-up questions before answering. \cite{press2023measuringnarrowingcompositionalitygap}
\end{tabular} \\ 
\midrule
Chain-of-Thought (CoT) & 1000 & 
\begin{tabular}[c]{@{}l@{}} 
- Let’s think step by step. \cite{cot}\\ 
- Break the question into subquestions. \cite{leasttomost}\\ 
- Take a deep breath and work on this problem step-by-step. \cite{opro} 
\end{tabular} \\ 
\midrule
Cross language & 1000 & 
\begin{tabular}[c]{@{}l@{}} 
- Prioritize responding in the language of the question. \\ 
- Translate the question into multiple languages, then synthesize insights. \\ 
- Use English for technical terms when necessary. \\ 
\end{tabular} \\ 
\bottomrule
\end{tabular}
}
\caption{List of prompt components in prompt component corpus.}
\label{tab:all_prompt_components}
\end{table*}

\subsection{Prompt Corpus Details}
\label{sec:corpus_details}
We generate 1,000 prompts in $\mathcal{P}$ by randomly combining prompt components from the corpus $\mathcal{C}$. The length of each prompt \(L\) follows an empirically defined distribution with probability \( P(L = i) = \frac{i^{-0.8}}{\sum_{j=1}^{30} j^{-0.8}} \), where \( i \in \{1, \dots, 30\} \), to ensure coverage across the full range of 0 to 30 components. %

\subsection{Reward Model Training}
\label{sec:reward_details}
Following prior work, we train our reward model using a max-margin pairwise loss~\cite{llama2} \( \mathcal{L}_{\text{ranking}} = -\log\left( \sigma\left( r_\theta(x, y_c) - r_\theta(x, y_r) - m(r) \right) \right) \), with ModernBERT~\cite{modernbert} as the backbone. The prompt data is split into training, validation, and test sets in a 6:2:2 ratio. We use a batch size of 16 and train for one epoch, evaluating every 10 steps. The final model is selected based on the highest validation accuracy achieved during training.

\subsection{Benchmark Translation Details}
\label{appendix:translation}
Although the three benchmarks (MMLU-Pro, MATH500, UniMoral) have multilingual variants online, we found during preliminary inspection that many existing Chinese and Hindi versions suffered from translation errors and encoding artifacts, particularly those generated by LLM-based pipelines. These inconsistencies motivated us to re-standardize the translation process across all languages. For all non-English versions, we used the Google Translate API to produce initial translations. To ensure quality and consistency, we conducted manual reviews with three proficient speakers of the five target languages, who each examined 50 randomly sampled examples. Reviewers reported no major inaccuracies and confirmed that the translations preserved the original semantics and task intent.

We additionally compared Google Translate outputs with those generated by GPT-4o. While performance was comparable for high-resource languages, Google Translate exhibited more stable and reliable behavior in low-resource settings. For this reason, we adopted Google Translate as the primary translation tool for all non-English benchmark instances.

\subsection{Multilingual SPRIG Pipeline}
We run all our experiments on 4 NVIDIA-L40S-48GB GPUs. All LLM inferences are powered by vLLM 0.5.4 \cite{kwon2023efficient}, Hugging Face Transformers 4.43.3 \cite{wolf-etal-2020-transformers} and PyTorch 2.4.0 \cite{NEURIPS2019_9015} on a CUDA 12.4 environment. Temperatures are set to 0.0 to minimize the effect of randomness.

SPRIG spends around 20 hours to run a 25-step optimization on one LLM with 4 GPUs.

\subsection{Reasoning Unit Segmentation}
\label{appendix:segmentation}
We segment LLM-generated reasoning into step-level units using a pretrained token classification model \textbf{appier-ai-research/reasoning-segmentation-model-v0} from the \href{https://github.com/appier-research/language-matters}{\textit{language-matters}}~\cite{language-matters} repository. To prepare inputs, we replace all line breaks in the reasoning output with \texttt{[SEP]} tokens. The model then predicts, for each \texttt{[SEP]}, whether it indicates a step boundary. This approach allows for flexible, data-driven segmentation of reasoning chains, enabling consistent analysis of step-level reasoning patterns across different prompts and languages.

\subsection{Reasoning Behavior Categorization}
\label{appendix:categorization_info}
We use Qwen3-14B-Instruct~\cite{qwen2.5} as the backbone of \textit{language-matters}~\cite{language-matters} pipeline to perform reasoning behavior classification throughout our main experiments. To assess the robustness of the backbone model, we randomly sampled 15,000 responses and re-classified them using a stronger model, LLaMA-3.1-70B-Instruct~\cite{llama3.1}, under the same pipeline. The results show a percentage agreement of 0.581 and a Cohen’s $\kappa$ score of 0.528 between the two models, indicating overall consistent annotations.

\subsection{Reasoning Category Justification}
\label{appendix:reasoning-categories}

The detailed definitions of the eight reasoning categories we use are listed below:

{\small\begin{lstlisting}
Subgoal setting: Where the model breaks down the problem into smaller, intermediate goals (e.g., 'To solve this, we first need to...' or 'First, I'll try to ..., then ...'
Backtracking: Where the model realizes a path won't work and explicitly goes back to try a different approach. An example of backtracking is: 'Let me try again' or 'we need to try a different approach'.
Verification: Where the model checks the correctness of the intermediate results or to make sure the final answer is correct.
Backward chaining: Where the model works backward from its answer to see whether it can derive the variables in the original problem.
Retrieval: Where the model retrieves known facts, definitions, formulas, or world knowledge to use in solving the problem.
Reframing: Where the model rephrases a question or problem to clarify its understanding or to approach it from a different angle. This includes paraphrasing, summarizing, or changing the perspective of the question.
Logical Reasoning: Where the model uses logical reasoning to arrive at the answer. This includes deductive reasoning, inductive reasoning, and other forms of logical inference.
Calculation: Where the model performs a calculation to arrive at the answer. This includes arithmetic operations, algebraic manipulations, or any other mathematical operations.
\end{lstlisting}
}

The first four reasoning categories are adopted from a recent paper~\cite{fourhabits}, while the remaining four are introduced based on an extensive literature review to address the limitations of the original taxonomy when applied to a broader range of tasks. The justification of newly defined categories are as follows:

\begin{enumerate}
    \item \textbf{Retrieval}: In a recent position paper on LLM reasoning, \citet{Kambhampati_2024} argues that what often appears to be reasoning in large language models is largely ``a form of universal approximate retrieval'' driven by their web-scale training data. Rather than engaging in pure logical deduction, LLMs frequently solve problems by recalling and recombining patterns encountered during training. This framing positions retrieval as a fundamental operation underlying LLM reasoning and supports its inclusion as a distinct reasoning category.

    \item \textbf{Reframing}: Cognitive science has long recognized reframing---recasting a problem from a different perspective---as a critical component of human reasoning~\cite{Bieth2024SemanticRestructuring}. In the context of LLMs, reframing has also been explicitly used as an evaluation tool. For example, \citet{reframingprompting} has shown that prompting LLM to reframing can be an empirically grounded strategy to boost performance, underscoring reframing as a meaningful reasoning strategy.

    \item \textbf{Logical Reasoning}: In reasoning-related research, Logical Reasoning is usually distinguished as a fundamental category of reasoning in LLMs, treated as explicit symbolic or deductive inference (rather than statistical pattern re‑combination). For example, recent work like Atomic Reasoner~\cite{liu2025chaosorderatomicreasoner} shows that modeling reasoning as modular logical units enhances coherence and control in complex problem-solving tasks, which supports logical reasoning as a distinct and essential category.

    \item \textbf{Calculation}: Numeric and arithmetic reasoning has frequently been treated as its own category in LLM evaluations. \citet{qiao-etal-2023-reasoning} explicitly list arithmetic reasoning as a distinct reasoning type, separate from commonsense or logical reasoning. Similarly, \citet{parashar2025inferencetimecomputationsllmreasoning} categorize LLM evaluation tasks into five primary reasoning types, including arithmetic reasoning as a standalone class. These consistent distinctions across domains support the inclusion of calculation as a meaningful and separable reasoning unit.
\end{enumerate}

\subsection{Language Identification}
\label{appendix:language-identification}
We use a pretrained language identification model from \textbf{fastText}~\cite{fasttext}, which supports prediction over 176 languages. The model was developed by Facebook AI Research and is widely used for multilingual text classification and language detection tasks. For each reasoning step, we apply the model to overlapping 100-character windows and retain language predictions with confidence above 0.6. All valid predictions are aggregated across windows to yield step-level language annotations, without applying heuristic preferences when multiple languages are detected.

\subsection{Full Experiment Results}

\begin{figure}[t]
  \includegraphics[width=\columnwidth]{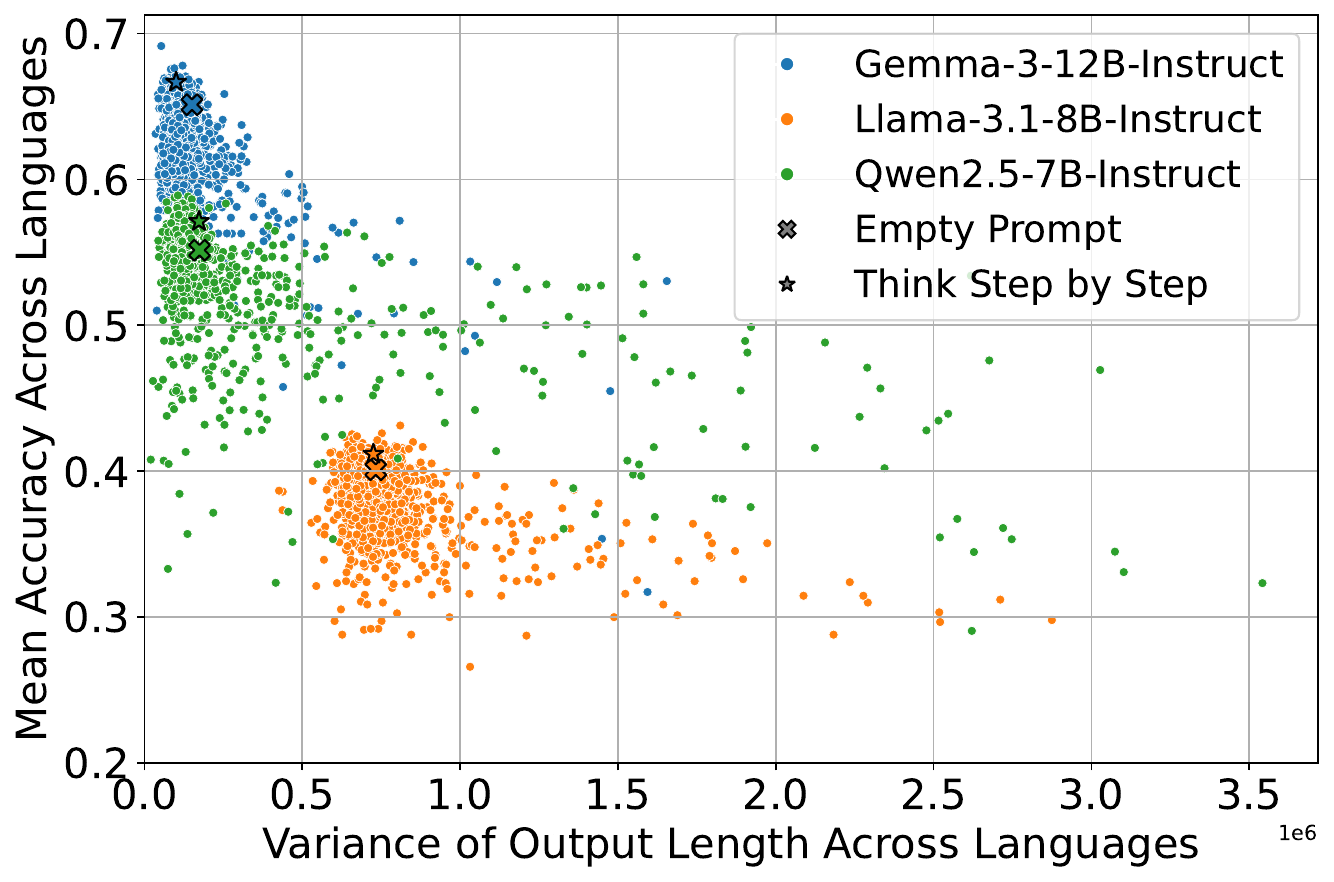}
  \caption{Relationship between mean accuracy ($\mathsf{Acc_{mean}}$) and output length variance ($\mathsf{len_{var}}$), with each point represents one prompt aggregated on all tasks.}
  \label{fig:Exp1-rq2-acc_mean_vs_len_var}
\end{figure}

Figure~\ref{fig:Exp1-rq2-acc_mean_vs_len_var} shows the relationship between mean accuracy ($\mathsf{Acc_{mean}}$) and output length variance ($\mathsf{len_{var}}$).

\begin{figure}[t]
  \includegraphics[width=\columnwidth]{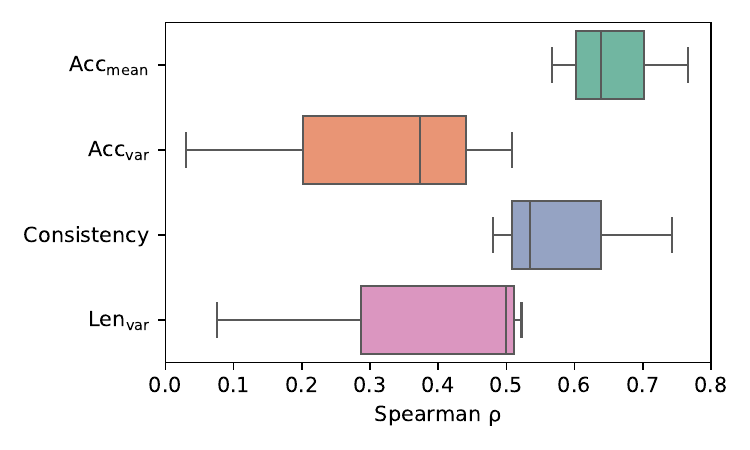}
  \caption{Reward model performance across four metrics. The model ranks $\mathsf{Acc_{mean}}$ and $\mathsf{Consistency}$ with high Spearman correlations, but correlations for $\mathsf{Acc_{var}}$ and $\mathsf{Len_{var}}$ are more variable.}
\label{fig:Exp2_sprig_reward_spearman}
\end{figure}

Figure~\ref{fig:Exp2_sprig_reward_spearman} shows the Spearman correlation of the trained system prompt reward model.

\begin{figure}[t]
  \includegraphics[width=\columnwidth]{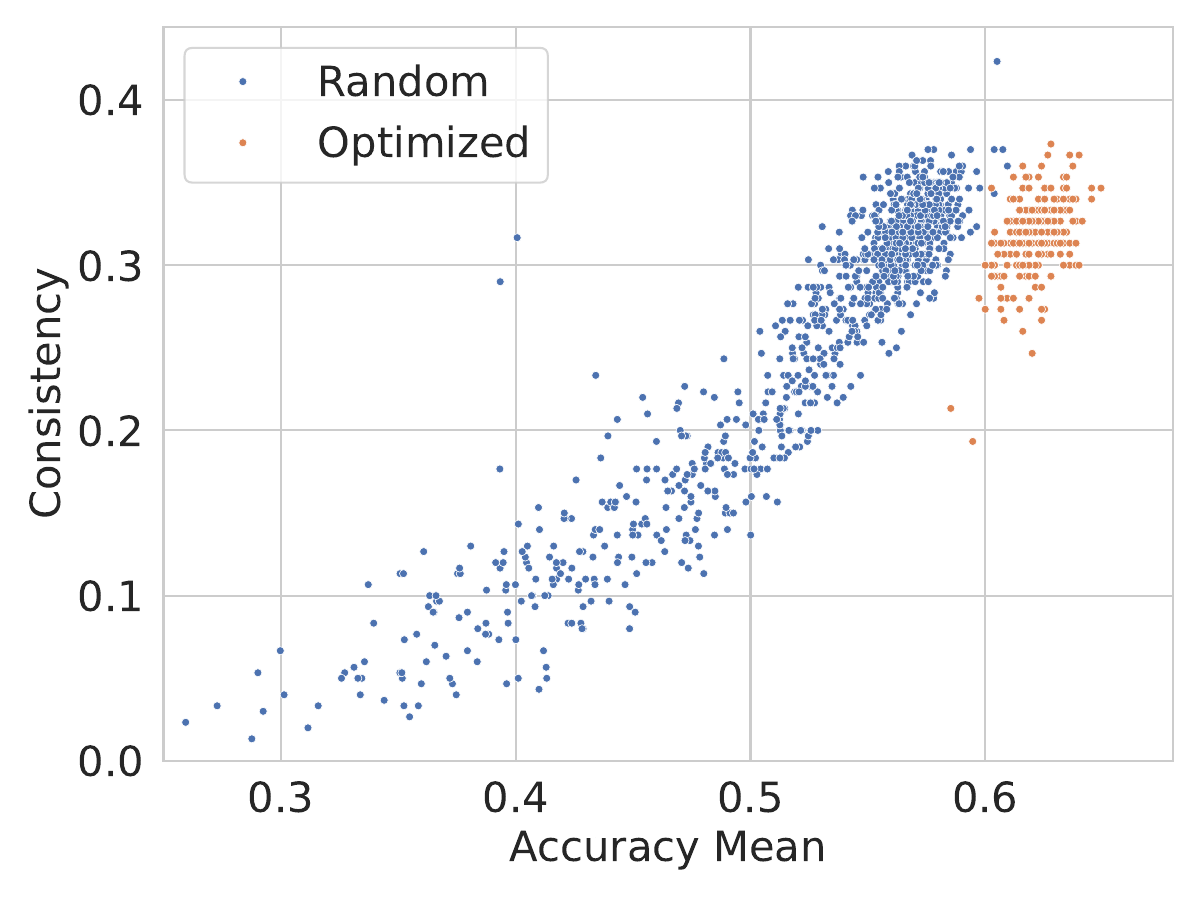}
  \caption{Scatter plot illustrating the distribution of accuracy mean and consistency for random versus optimized prompts.}
  \label{fig:Exp2_acc_mean_vs_consistency}
\end{figure}

\begin{figure}[t]
  \includegraphics[width=\columnwidth]{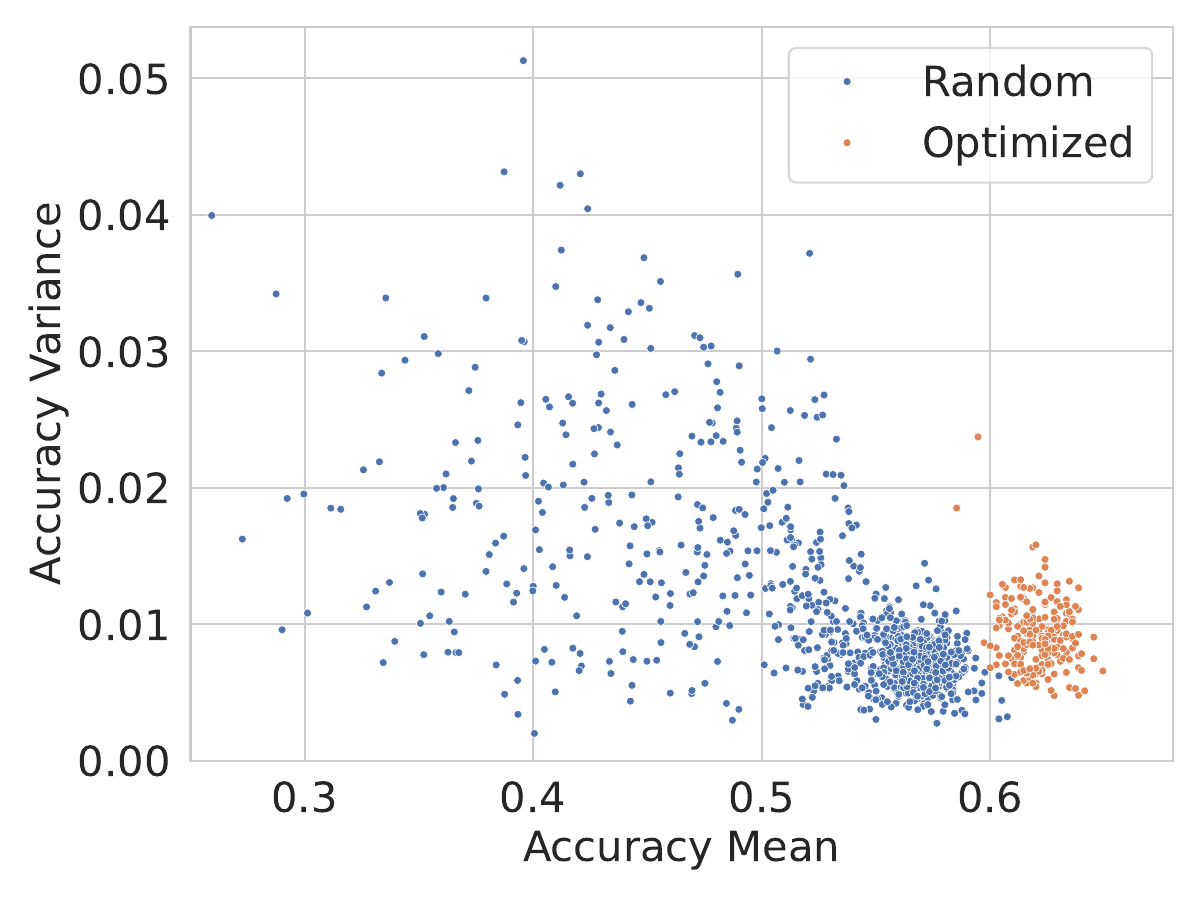}
  \caption{Scatter plot illustrating the distribution of accuracy mean and accuracy variance for random versus optimized prompts.}
  \label{fig:Exp2_acc_mean_vs_acc_var}
\end{figure}

In Figure \ref{fig:Exp2_acc_mean_vs_consistency}, \ref{fig:Exp2_acc_mean_vs_acc_var}, we illustrate how Consistency and Accuracy Variance varies after optimization.

\begin{figure}[t]
  \includegraphics[width=\columnwidth]{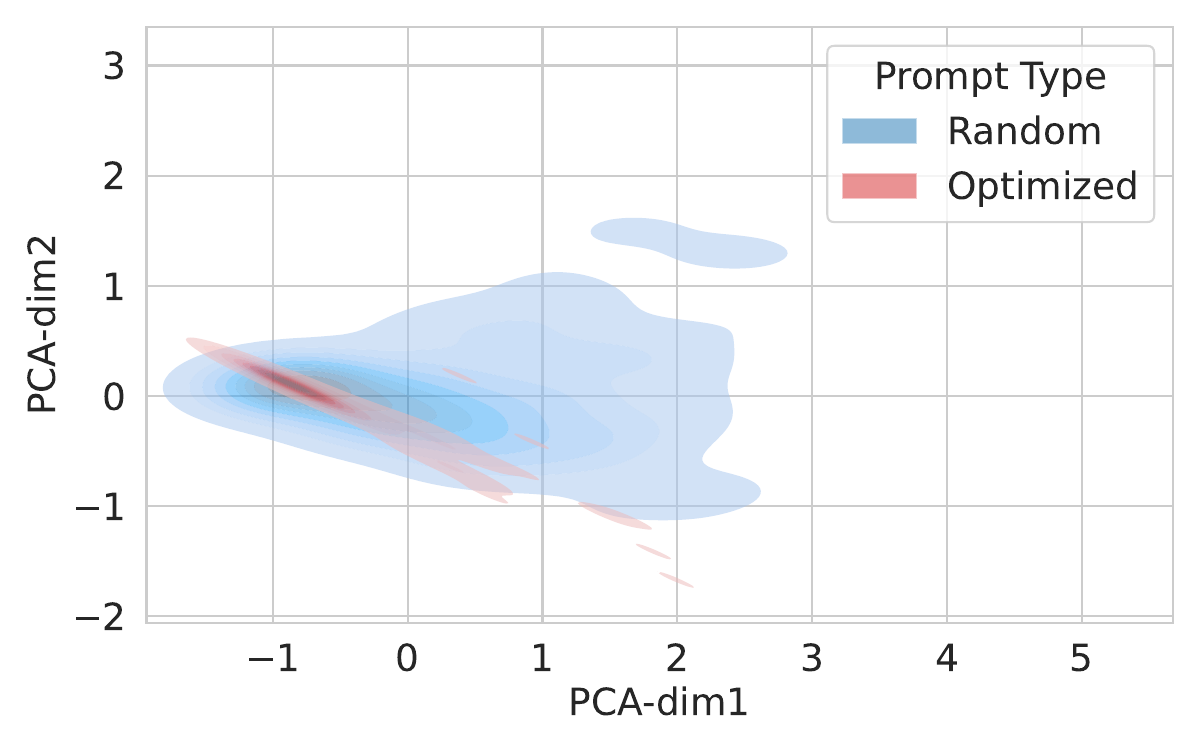}
  \caption{KDE plot showing the distribution of prompts. This figure is based on results from Qwen2.5-7B-Instruct on the Unimoral
benchmark}
  \label{fig:Exp3_pca_unimoral}
\end{figure}

\begin{figure}[t]
  \includegraphics[width=\columnwidth]{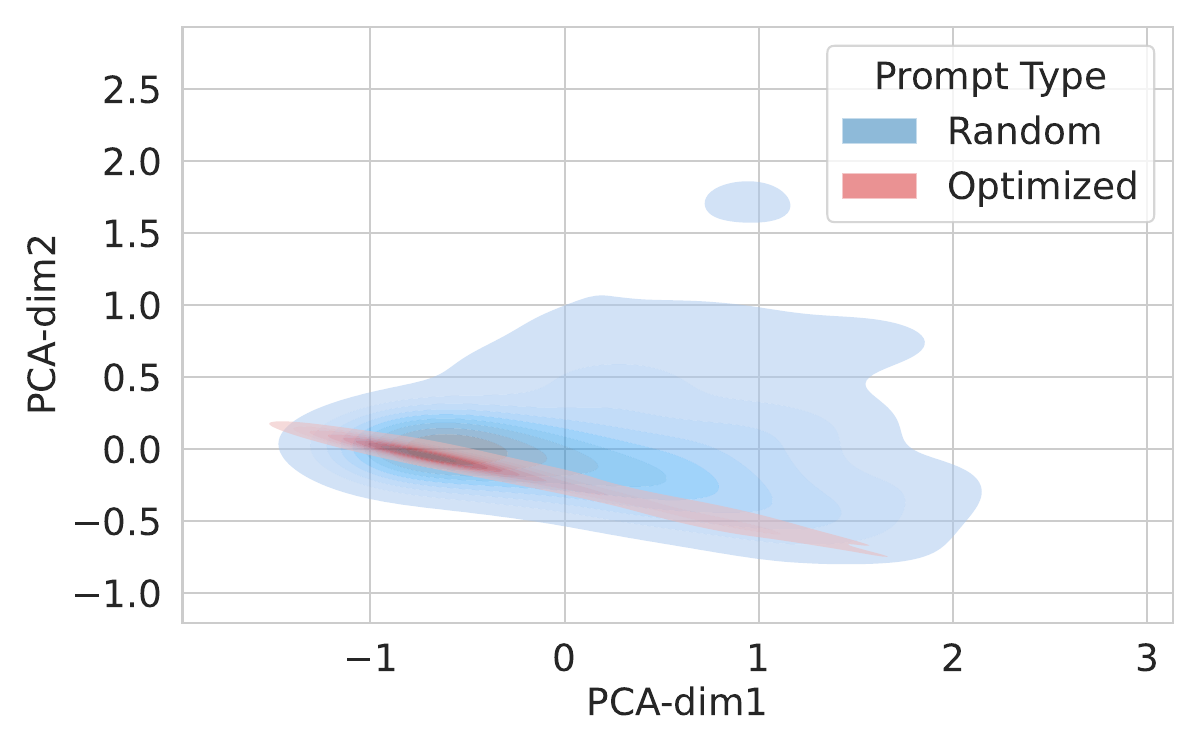}
  \caption{KDE plot showing the distribution of prompts. This figure is based on results from Qwen2.5-7B-Instruct on the MMLU-Pro
benchmark}
  \label{fig:Exp3_pca_mmlupro}
\end{figure}

Figure~\ref{fig:Exp3_pca_unimoral} and Figure~\ref{fig:Exp3_pca_mmlupro} shows the KDE plot before and after optimization on benchmark \textbf{Unimoral} and \textbf{MMLU-Pro}.

\begin{figure*}[tb!]
    \centering
    \includegraphics[width=\textwidth]{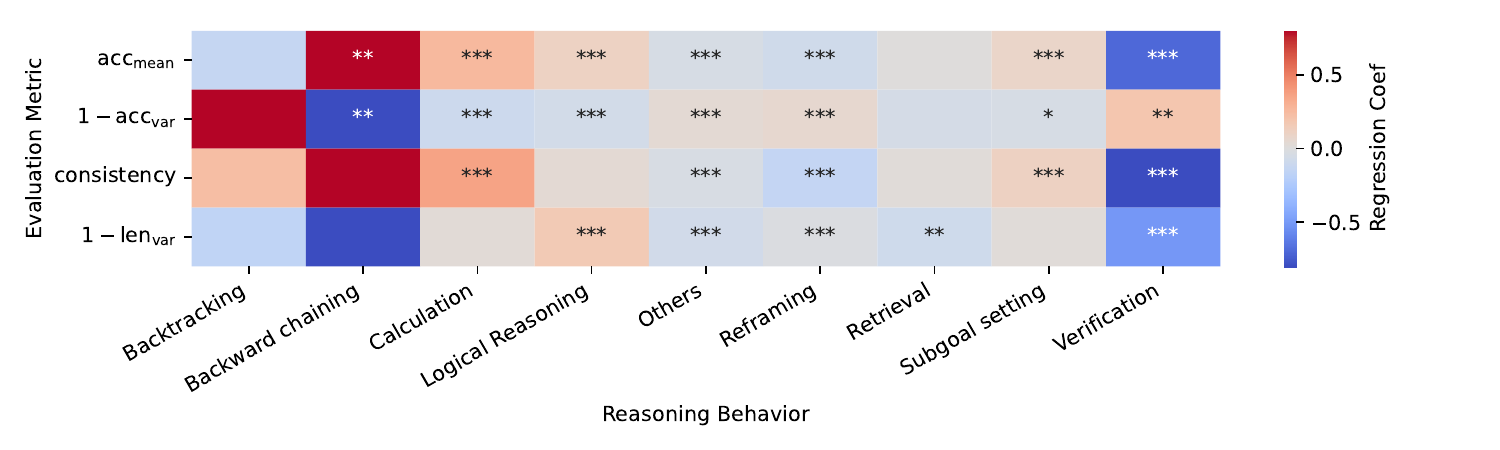}
        \caption{The regression heatmap illustrating the impact of different reasoning behaviors on multilingual model behavior (* p<0.05, ** p<0.01, *** p<0.001). Results are averaged across benchmarks.}
    \label{fig:Exp3-regression}
\end{figure*}

\begin{figure*}[tb!]
    \centering
    \includegraphics[width=\textwidth]{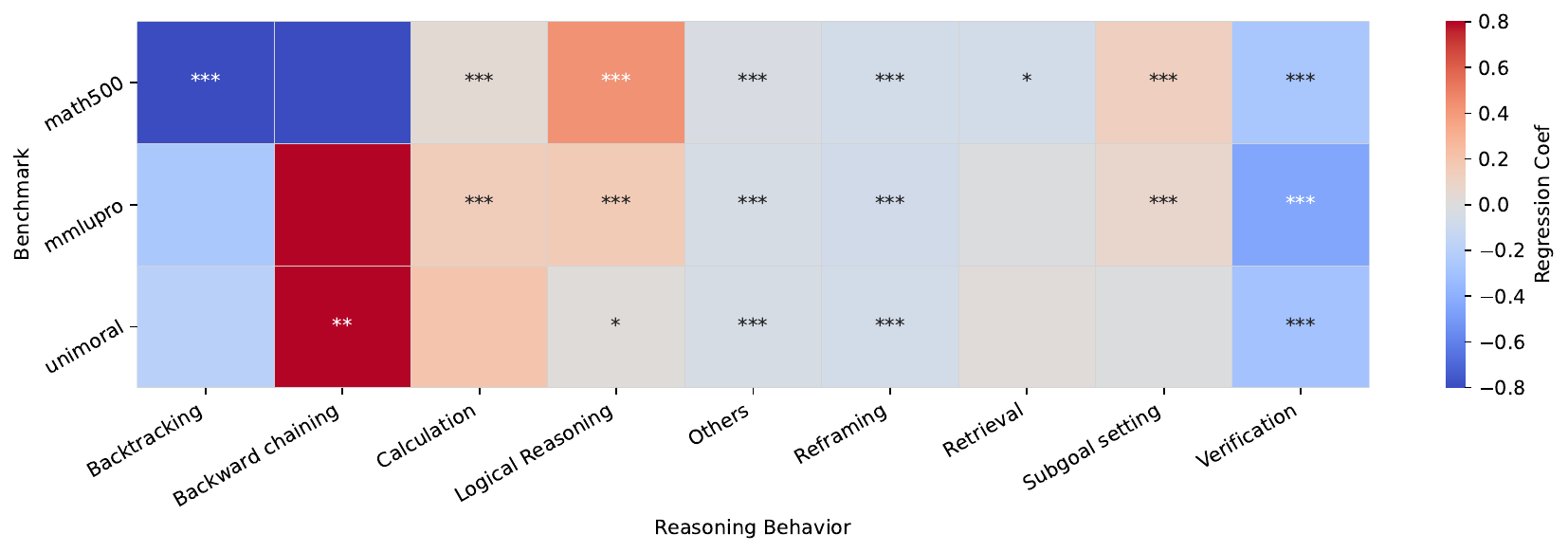}
        \caption{Regression results of reasoning behaviors on accuracy mean across different benchmarks.}
    \label{fig:Exp3-regression_avgacc}
\end{figure*}

\begin{figure*}[tb!]
    \centering
    \includegraphics[width=\textwidth]{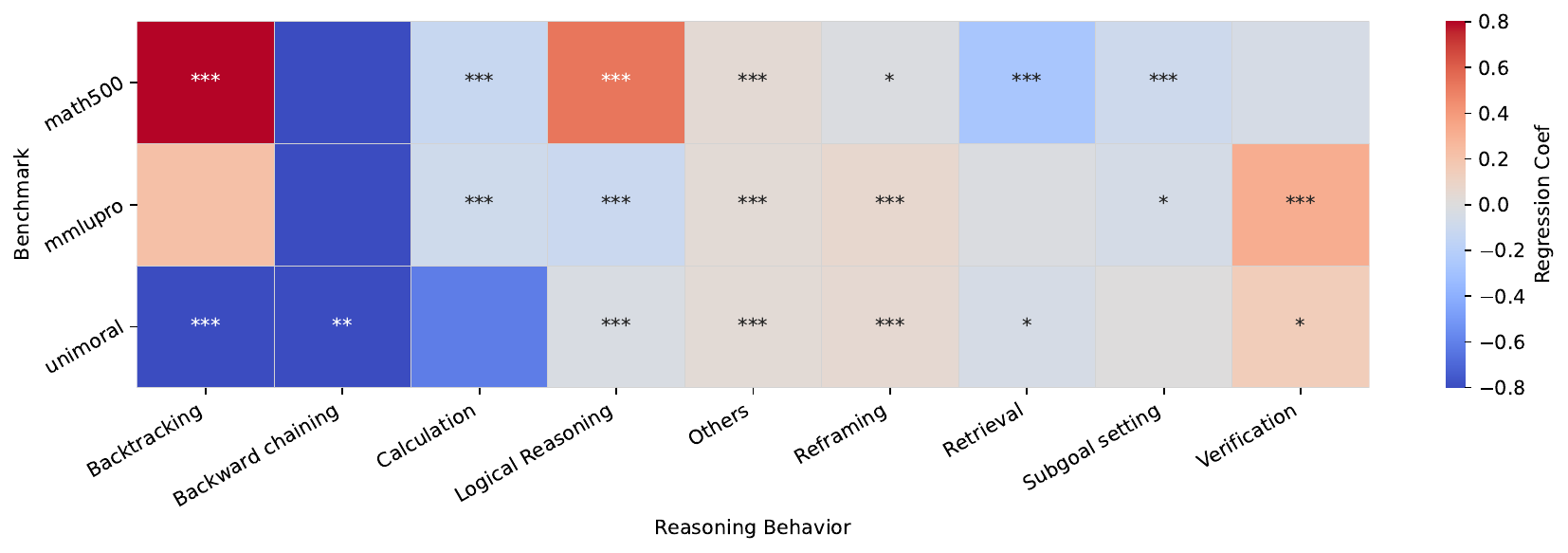}
        \caption{Regression results of reasoning behaviors on accuracy variance across different benchmarks.}
    \label{fig:Exp3-regression_accvar}
\end{figure*}

\begin{figure*}[tb!]
    \centering
    \includegraphics[width=\textwidth]{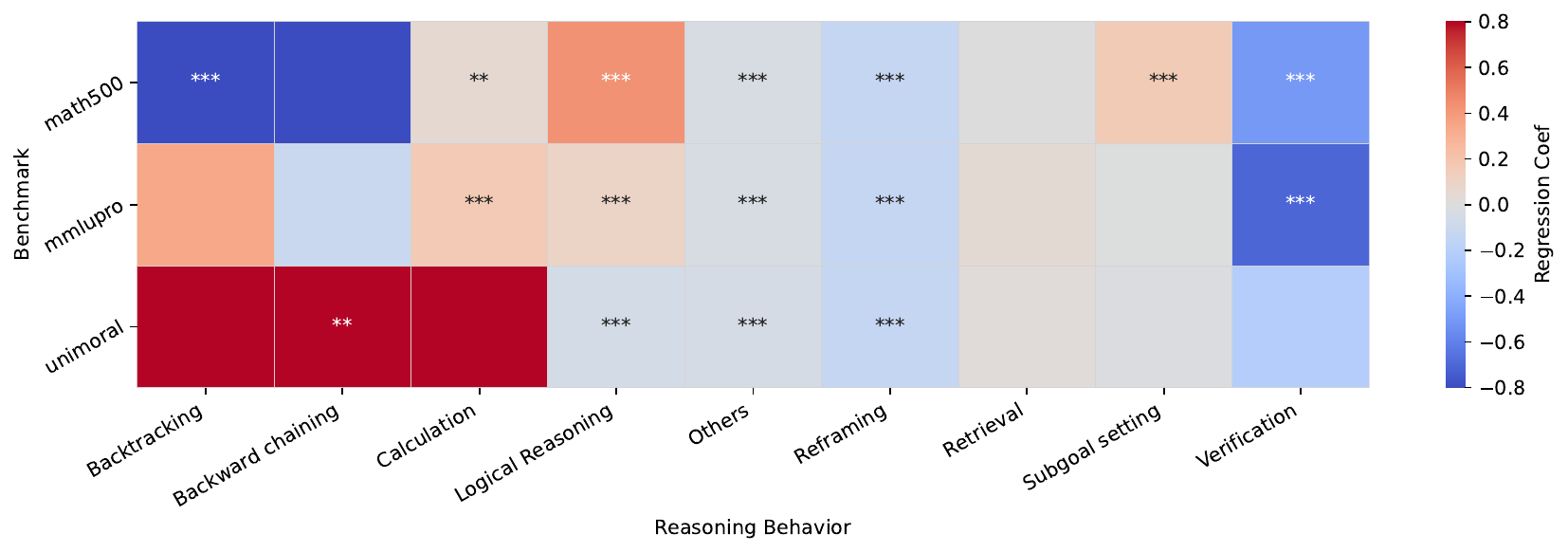}
        \caption{Regression results of reasoning behaviors on consistency across different benchmarks.}
    \label{fig:Exp3-regression_consistency}
\end{figure*}

\begin{figure*}[tb!]
    \centering
    \includegraphics[width=\textwidth]{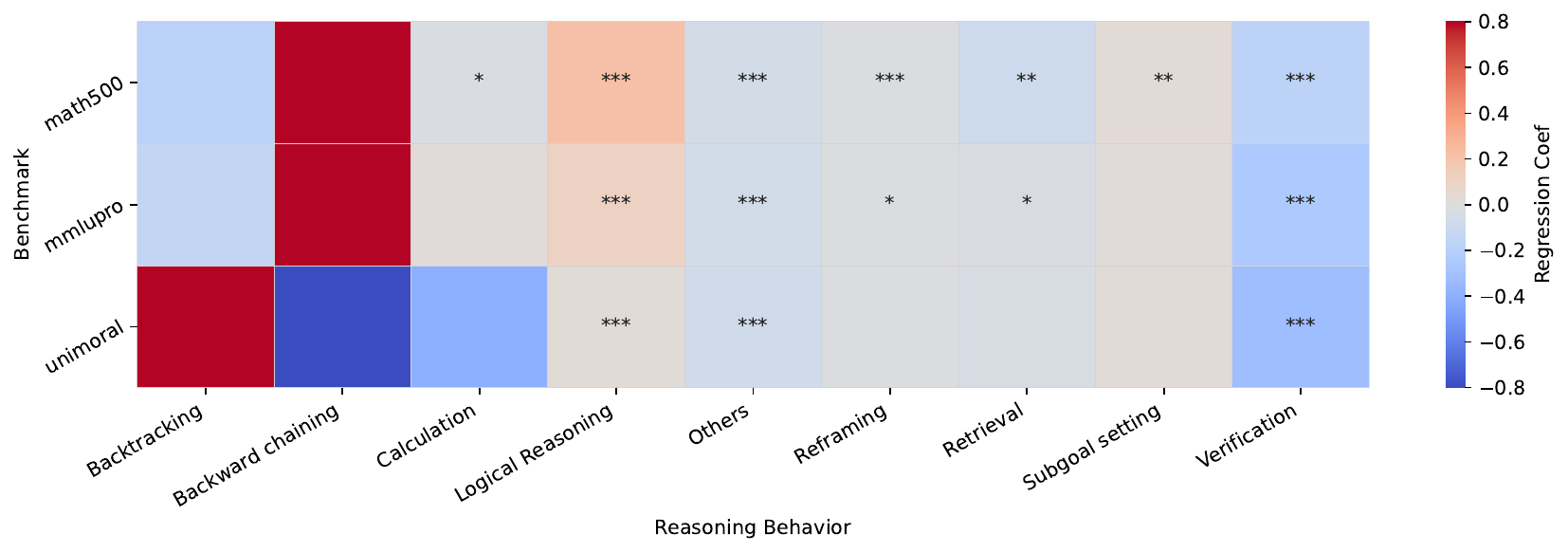}
        \caption{Regression results of reasoning behaviors on output length variance across different benchmarks.}
    \label{fig:Exp3-regression_length_var}
\end{figure*}

Figure~\ref{fig:Exp3-regression} shows the combined regression heatmap illustrating the impact of different reasoning behaviors on multilingual model behavior.

In Figure~\ref{fig:Exp3-regression_avgacc}, \ref{fig:Exp3-regression_accvar}, \ref{fig:Exp3-regression_consistency}, and \ref{fig:Exp3-regression_length_var}, we show the regression results of reasoning behaviors on the four evaluation metrics across different benchmarks.

\subsection{Full Experiment Results(table)}
Table \ref{tab:optimizer_results_all} summarizes the overall performance of prompts on Qwen2.5-7B-Instruct, Llama-3.1-8B-Instruct and Gemma-3-12B-Instruct.

Table \ref{tab:spearman_qwen_sizes} shows the Spearman correlation of the prompt-ranking across Qwen2.5 model family of different sizes.

In Table \ref{tab:rq1_1_metrics}, we presented the detailed data for overall performance of tow different prompt settings.

Table \ref{tab:rq1_3_regression} summarize the regression results of different component types across various metrics.

Table \ref{tab:lang_shift_before_after_absolute} and \ref{tab:lang_shift_before_after_percent} respectively show the absolute counts and relative proportions of different languages in LLM responses before and after optimization. 

In Table \ref{tab:reasoning_behavior_absolute} and \ref{tab:reasoning_behavior_proportion}, we respectively presented the absolute counts and proportion of reasoning behaviors across different response language.

Table \ref{tab:reasoning_counts_grouped} shows the average counts of reasoning behavior by benchmark and prompt type.

\subsection{Raw Data}

\begin{table}[ht]
\centering
\small
\begin{tabular}{lcc}
\toprule
\textbf{Metric} & \textbf{English Prompt} & \textbf{Same Language} \\
\midrule
Acc\_mean & 0.5033 & 0.5097 \\
Acc\_var & 0.0108 & 0.0100 \\
Consistency & 0.2653 & 0.2676 \\
Output\_tokens\_var & 431338 & 414644 \\
\bottomrule
\end{tabular}
\caption{Comparison of performance metrics between English and same-language prompts.}
\label{tab:rq1_1_metrics}
\end{table}

\begin{table*}[t]
\centering
\small
\begin{tabular}{lcccccccc}
\toprule
\textbf{Component} & \multicolumn{2}{c}{\textbf{Acc\_mean}} & \multicolumn{2}{c}{\textbf{Acc\_var}} & \multicolumn{2}{c}{\textbf{Consistency}} & \multicolumn{2}{c}{\textbf{Len\_var}} \\
\cmidrule(lr){2-3} \cmidrule(lr){4-5} \cmidrule(lr){6-7} \cmidrule(lr){8-9}
 & \textbf{$k$} & \textbf{$p$} & \textbf{$k$} & \textbf{$p$} & \textbf{$k$} & \textbf{$p$} & \textbf{$k$} & \textbf{$p$} \\
\midrule
CoT & 0.0264 & 4.37e-08 & 0.0303 & 1.52e-05 & 0.0292 & 9.36e-06 & -0.0136 & 0.0093 \\
behavioral & -0.0209 & 1.44e-05 & 0.0174 & 0.0127 & -0.0269 & 4.57e-05 & -0.0087 & 0.0963 \\
cross-language & -0.1216 & 6.51e-128 & -0.0751 & 1.69e-27 & -0.1684 & 2.35e-130 & -0.0869 & 1.72e-60 \\
emotion & 0.0207 & 1.26e-05 & 0.0215 & 0.0018 & 0.0316 & 1.16e-06 & 0.0174 & 0.0007 \\
good\_property & 0.0058 & 0.2306 & -0.0102 & 0.1448 & -0.0009 & 0.8955 & 0.0075 & 0.1541 \\
jailbreak & 0.0066 & 0.1735 & 0.0196 & 0.0053 & 0.0037 & 0.5819 & -0.0016 & 0.7559 \\
role & -0.0151 & 0.0020 & -0.0577 & 5.43e-16 & -0.0295 & 1.02e-05 & -0.0006 & 0.9036 \\
safety & -0.0014 & 0.7729 & -0.0002 & 0.9768 & 0.0061 & 0.3551 & 0.0140 & 0.0084 \\
scenario & 0.0122 & 0.0102 & 0.0301 & 1.23e-05 & 0.0202 & 0.0019 & -0.0028 & 0.5815 \\
style & -0.0473 & 2.60e-23 & -0.0361 & 1.31e-07 & -0.0488 & 4.54e-14 & -0.0135 & 0.0084 \\
\bottomrule
\end{tabular}
\caption{Regression coefficients ($k$) and $p$-values for each component across four evaluation metrics.}
\label{tab:rq1_3_regression}
\end{table*}

\begin{table*}[ht]
\centering
\small
\begin{tabular}{lllcccc}
\toprule
Model & Benchmark & Setting & Acc\_mean & Acc\_var & Consistency & Output\_tokens\_var \\
\midrule
\multirow{7}{*}{Qwen2.5-7B-Instruct} 
& MMLU-Pro & Random    & 0.399 & 0.015 & 0.114 & 372446.18 \\
&          & Optimized & 0.497 & 0.018 & 0.174 & 128362.80 \\
& UNIMORAL & Random    & 0.577 & 0.011 & 0.295 & 354881.31 \\
&          & Optimized & 0.679 & 0.003 & 0.405 & 20826.80  \\
& MATH500  & Random    & 0.585 & 0.007 & 0.354 & 305133.92 \\
&          & Optimized & 0.686 & 0.007 & 0.354 & 122866.70 \\
& Mean     & Random    & 0.521 & 0.011 & 0.254 & 344153.80 \\
&          & Optimized & 0.621 & 0.009 & 0.311 & 90418.77  \\
\midrule
\multirow{7}{*}{Llama-3.1-8B-Instruct} 
& MMLU-Pro & Random    & 0.319 & 0.010 & 0.109 & 822103.05 \\
&          & Optimized & 0.373 & 0.011 & 0.155 & 513244.20 \\
& UNIMORAL & Random    & 0.559 & 0.024 & 0.240 & 171978.54 \\
&          & Optimized & 0.589 & 0.004 & 0.473 & 13603.50  \\
& MATH500  & Random    & 0.249 & 0.015 & 0.027 & 1512173.79 \\
&          & Optimized & 0.285 & 0.026 & 0.011 & 1547849.00 \\
& Mean     & Random    & 0.376 & 0.016 & 0.125 & 835418.46 \\
&          & Optimized & 0.416 & 0.014 & 0.213 & 692565.57 \\
\midrule
\multirow{7}{*}{Gemma-3-12B-Instruct}
& MMLU-Pro & Random    & 0.501 & 0.008  & 0.263 & 114524.53 \\
&          & Optimized & 0.656 & 0.0128 & 0.333 & 102069.60 \\
& UNIMORAL & Random    & 0.669 & 0.005  & 0.491 & 75661.40  \\
&          & Optimized & 0.640 & 0.0018 & 0.498 & 49311.87  \\
& MATH500  & Random    & 0.670 & 0.002  & 0.495 & 153144.84 \\
&          & Optimized & 0.797 & 0.0016 & 0.554 & 144471.24 \\
& Mean     & Random    & 0.614 & 0.005  & 0.416 & 114443.59 \\
&          & Optimized & 0.698 & 0.0054 & 0.462 & 98617.57  \\

\bottomrule
\end{tabular}
\caption{
Performance of optimized vs. random prompts across three benchmarks for Qwen2.5-7B-Instruct, Llama-3.1-8B-Instruct, and Gemma-3-12B-Instruct.
}
\label{tab:optimizer_results_all}
\end{table*}

\subsection{Scaling Law}
\begin{table*}[ht]
\centering
\small
\begin{tabular}{lccccc}
\toprule
& Qwen2.5-1.5B & Qwen2.5-3B & Qwen2.5-7B & Qwen2.5-14B & Qwen2.5-32B \\
\midrule
Qwen2.5-1.5B & 1.00 & 0.31 & 0.28 & 0.26 & 0.25 \\
Qwen2.5-3B   & 0.31 & 1.00 & 0.68 & 0.61 & 0.53 \\
Qwen2.5-7B   & 0.28 & 0.68 & 1.00 & 0.70 & 0.56 \\
Qwen2.5-14B  & 0.26 & 0.61 & 0.70 & 1.00 & 0.69 \\
Qwen2.5-32B  & 0.25 & 0.53 & 0.56 & 0.69 & 1.00 \\
\bottomrule
\end{tabular}
\caption{Spearman correlation of prompt-ranking across Qwen2.5 model family.}
\label{tab:spearman_qwen_sizes}
\end{table*}

\subsection{Detailed Data}

\begin{table*}[t]
\centering
\small
\begin{tabular}{lcccccc}
\toprule
\multirow{2}{*}{\textbf{Question Language}} & \multicolumn{3}{c}{\textbf{Before}} & \multicolumn{3}{c}{\textbf{After}} \\
\cmidrule(lr){2-4} \cmidrule(lr){5-7}
 & \textbf{Question\_lang} & \textbf{En} & \textbf{Other} & \textbf{Question\_lang} & \textbf{En} & \textbf{Other} \\
\midrule
English & 4.27 & 0.00 & 1.81 & 3.67 & 0.00 & 0.23 \\
Spanish & 2.50 & 0.98 & 1.55 & 2.83 & 0.07 & 0.07 \\
French  & 2.44 & 1.04 & 2.05 & 3.06 & 0.06 & 0.07 \\
Hindi   & 2.49 & 1.62 & 2.02 & 2.87 & 0.82 & 0.34 \\
Chinese & 3.54 & 0.79 & 1.52 & 3.98 & 0.01 & 0.03 \\
\bottomrule
\end{tabular}
\caption{Language usage of model outputs before and after optimization across task languages. Each cell shows the absolute number of outputs in the question language, English, or other languages.}
\label{tab:lang_shift_before_after_absolute}
\end{table*}

\begin{table*}[t]
\centering
\small
\begin{tabular}{lcccccc}
\toprule
\multirow{2}{*}{\textbf{Question Language}} & \multicolumn{3}{c}{\textbf{Before}} & \multicolumn{3}{c}{\textbf{After}} \\
\cmidrule(lr){2-4} \cmidrule(lr){5-7}
 & \textbf{Question\_lang (\%)} & \textbf{En (\%)} & \textbf{Other (\%)} & \textbf{Question\_lang (\%)} & \textbf{En (\%)} & \textbf{Other (\%)} \\
\midrule
English & 70.25 & 0.00 & 29.75 & 94.08 & 0.00 & 5.92 \\
Spanish & 49.68 & 19.41 & 30.91 & 95.29 & 2.25 & 2.46 \\
French  & 44.15 & 18.76 & 37.09 & 95.73 & 2.02 & 2.25 \\
Hindi   & 40.60 & 26.38 & 33.02 & 71.27 & 20.26 & 8.47 \\
Chinese & 60.52 & 13.54 & 25.95 & 98.91 & 0.33 & 0.77 \\
\bottomrule
\end{tabular}
\caption{Proportion (\%) of model outputs by language category before and after optimization. Each cell shows the share of outputs in the task language, English, or other languages.}
\label{tab:lang_shift_before_after_percent}
\end{table*}

\begin{table*}[t]
\centering
\small
\begin{tabular}{lccccc}
\toprule
\textbf{Reasoning Behavior} & \textbf{English} & \textbf{Spanish} & \textbf{French} & \textbf{Hindi} & \textbf{Chinese} \\
\midrule
Retrieval & 1{,}075{,}338 & 337{,}538 & 300{,}245 & 278{,}707 & 462{,}074 \\
Reframing & 1{,}313{,}931 & 276{,}343 & 259{,}302 & 287{,}835 & 355{,}964 \\
Logical Reasoning & 1{,}284{,}127 & 348{,}258 & 350{,}576 & 312{,}341 & 577{,}093 \\
Calculation & 1{,}746{,}134 & 534{,}976 & 480{,}666 & 371{,}627 & 556{,}491 \\
Subgoal Setting & 781{,}733 & 191{,}862 & 202{,}660 & 172{,}775 & 257{,}763 \\
Backtracking & 18{,}689 & 7{,}889 & 6{,}624 & 6{,}887 & 8{,}270 \\
Verification & 331{,}654 & 116{,}351 & 94{,}897 & 102{,}664 & 127{,}305 \\
Backward Chaining & 544 & 153 & 153 & 234 & 317 \\
Others & 1{,}615{,}502 & 410{,}050 & 395{,}965 & 708{,}411 & 506{,}225 \\
\bottomrule
\end{tabular}
\caption{Absolute counts of reasoning behaviors across languages.}
\label{tab:reasoning_behavior_absolute}
\end{table*}

\begin{table*}[t]
\centering
\small
\begin{tabular}{lccccc}
\toprule
\textbf{Reasoning Behavior} & \textbf{English} & \textbf{Spanish} & \textbf{French} & \textbf{Hindi} & \textbf{Chinese} \\
\midrule
Retrieval & 0.132 & 0.152 & 0.144 & 0.124 & 0.162 \\
Reframing & 0.161 & 0.124 & 0.124 & 0.128 & 0.125 \\
Logical Reasoning & 0.157 & 0.157 & 0.168 & 0.139 & 0.202 \\
Calculation & 0.214 & 0.241 & 0.230 & 0.166 & 0.195 \\
Subgoal Setting & 0.096 & 0.086 & 0.097 & 0.077 & 0.090 \\
Backtracking & 0.002 & 0.004 & 0.003 & 0.003 & 0.003 \\
Verification & 0.041 & 0.052 & 0.045 & 0.046 & 0.045 \\
Backward Chaining & 0.00007 & 0.00007 & 0.00007 & 0.00010 & 0.00011 \\
Others & 0.198 & 0.184 & 0.189 & 0.316 & 0.178 \\
\bottomrule
\end{tabular}
\caption{Proportion of reasoning behaviors across languages.}
\label{tab:reasoning_behavior_proportion}
\end{table*}

\begin{table*}[t]
\centering
\small
\begin{tabular}{llccc}
\toprule
\textbf{Reasoning Type} & \textbf{Type} & \textbf{math500} & \textbf{mmlupro} & \textbf{unimoral} \\
\midrule
Backtracking & Randomized & 0.0146 & 0.0199 & 0.0020 \\
             & Optimized  & 0.0233 & 0.0196 & 0.0007 \\
Backward chaining & Randomized & 0.00045 & 0.00064 & 0.00058 \\
                  & Optimized  & 0.00025 & 0.00056 & 0.00033 \\
Retrieval & Randomized & 0.9920 & 1.3743 & 0.1869 \\
          & Optimized  & 1.0472 & 1.1494 & 0.0740 \\
Verification & Randomized & 0.2179 & 0.4402 & 0.0910 \\
             & Optimized  & 0.2104 & 0.2698 & 0.0110 \\
Subgoal setting & Randomized & 0.8353 & 0.4742 & 0.2999 \\
                & Optimized  & 0.8416 & 0.3161 & 0.0522 \\
Reframing & Randomized & 1.0096 & 0.8734 & 1.0244 \\
          & Optimized  & 0.6960 & 0.1803 & 0.0587 \\
Others & Randomized & 1.4342 & 1.4528 & 1.6386 \\
       & Optimized  & 0.7082 & 0.6355 & 0.9067 \\
Calculation & Randomized & 2.7463 & 0.9178 & 0.0004 \\
            & Optimized  & 2.8493 & 0.8161 & 0.0000 \\
Logical Reasoning & Randomized & 0.9130 & 0.9800 & 1.1505 \\
                     & Optimized  & 1.1287 & 0.7531 & 0.4664 \\
\bottomrule
\end{tabular}
\caption{Average counts of reasoning behaviors by benchmark and prompt type.}
\label{tab:reasoning_counts_grouped}
\end{table*}

\subsection{Licenses}
All data and code will be publicly released under the CC BY-SA 4.0 license.

\end{document}